\documentclass{bmvc2k}

\usepackage{booktabs}
\usepackage[ruled,vlined]{algorithm2e}
\usepackage{graphicx,subfigure}
\usepackage{tabularx}
\usepackage[dvipsnames]{xcolor}
\usepackage{enumitem}
\usepackage{geometry}

\title{Learning to Adapt Multi-View Stereo by Self-Supervision}
\addauthor{Arijit Mallick}{arijit.mallick@uni-tuebingen.de}{1}
\addauthor{J\"org St\"uckler}{joerg.stueckler@tuebingen.mpg.de}{2}
\addauthor{Hendrik Lensch}{hendrik.lensch@uni-tuebingen.de}{1}

\addinstitution{
 Computer Graphics Group\\
 University of T\"ubingen\\
 T\"ubingen, Germany
}
\addinstitution{
 Embodied Vision Group\\
 Max Planck Institute for Intelligent Systems\\
 T\"ubingen, Germany
}

\runninghead{Mallick et al.}{Adaptive Learning of Self-supervised Multi-View Stereo}

\begin{document}

\newgeometry{right=0.5mm} 
\maketitle

\begin{abstract}
3D scene reconstruction from multiple views is an important classical problem in computer vision. Deep learning based approaches have recently demonstrated impressive reconstruction results. When training such models, self-supervised methods are favourable since they do not rely on ground truth data which would be needed for supervised training and is often difficult to obtain. Moreover, learned multi-view stereo reconstruction is prone to environment changes and should robustly generalise to different domains. We propose an adaptive learning approach for multi-view stereo which trains a deep neural network for improved adaptability to new target domains. We use model-agnostic meta-learning (MAML) to train base parameters which, in turn, are adapted for multi-view stereo on new domains through self-supervised training. Our evaluations demonstrate that the proposed adaptation method is effective in learning self-supervised multi-view stereo reconstruction in new domains.
\end{abstract}
\restoregeometry

\section{Introduction}
\label{sec:intro}
Dense 3D scene reconstruction based on images from multiple view points is one of the classical challenges in computer vision.
It has widespread applications in areas such as computer aided design (CAD), virtual tours, augmented reality, cultural heritage preservation, construction maintenance and inspection, or robotics.
Given the known view poses and camera intrinsics, multi-view geometry is typically used to find correspondences between pixels of reference along epipolar lines.
Early approaches use handcrafted similarity measures for pixels or patches such as photometric similarity or normalized cross correlation.
Deep learning has recently been demonstrated as a capable alternative for learning image features from data which can excel handcrafted measures~\cite{Paschalidou_2018_CVPR,MVSNet,im2019dpsnet,huang2018deepmvs,Surfacenet,Ummenhofer_2017}.\\

\begin{figure}
	\centering     
	\subfigure[Buddha scan]{\label{fig:motivation_a}\includegraphics[width=25mm]{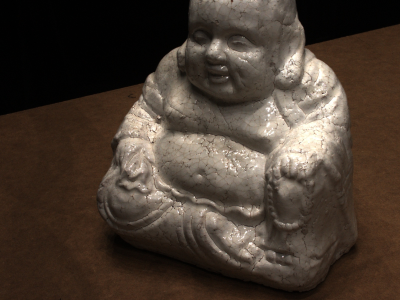}}
	\subfigure[Fine tuning]{\label{fig:motivation_b}\includegraphics[width=25mm]{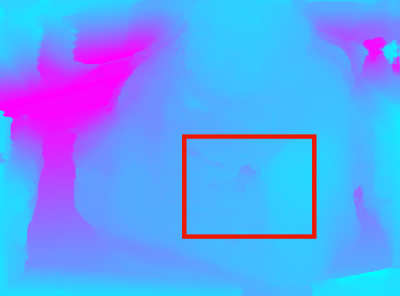}}
	\subfigure[Ours]{\label{fig:motivation_c}\includegraphics[width=25mm]{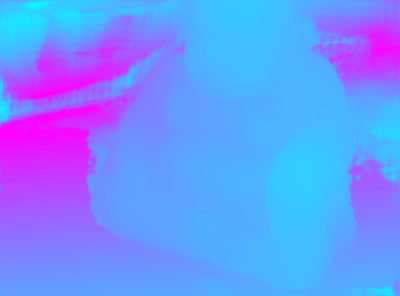}}
	\caption{Example result for adaptive meta-learning for self-supervised domain transfer. a) 3D scan (from DTU dataset), b) Reconstruction result for pre-training on BlendedMVS dataset without meta-learning and fine-tuning on DTU training set. c) Reconstruction result for our approach with meta-learning on BlendedMVS and self-supervised fine-tuning on DTU. Note the depth artifacts in the red box by the naive fine-tuning approach which do not occur in our meta-learning approach.}
	\label{fig:motivation}
\end{figure}
The state-of-the-art deep learning based methods for multi-view stereo reconstruction are supervised learning approaches which require immensive amounts of ground-truth 3D reconstruction data.
Yet such data is tedious and difficult to obtain.
Existing datasets such as \cite{DTU1,DTU2,sun3d} lack data diversity, come with calibration artifacts between the camera and the depth measuring device, or are synthetic.
Hence, self-supervised learning methods which can leverage large collections of camera images without the need of ground-truth 3D annotations are preferable. 
Apart from this, the given algorithm also needs to be robust against changes in environment or domains as it is not always possible to train a network with all possible environments in the training data. 
Hence, there needs to be a learning mechanism which can compensate for the changes in environment and quickly learn to adapt to different domains (indoors vs outdoors, low light vs bright light, building architecture scans vs object scans).
A motivating example in our context is highlighted in Fig.~\ref{fig:motivation}. 
Recent developments in meta learning~\cite{MAML} demonstrated online adaptation to new tasks of supervised regression models which have been trained on a different set of tasks.
In our approach, we propose a variant of model-agnostic meta-learning (MAML~\cite{MAML}) for training a multi-view stereo reconstruction network which facilitates self-supervised adaptation to new domains. 
We base our method on classical concepts from multi-view stereo (MVS) reconstruction and estimate dense depth in a reference view. 
Our model extends the network architecture of MVSNet~\cite{MVSNet} which has been demonstrated to yield state-of-the-art performance for supervised and self-supervised learning.
In a first training stage, we use our meta-learning approach to train a network on a large dataset in several domains with ground-truth depth annotation.
We train the network in such a way that it can better adapt to new domains through self-supervised training on data without ground-truth depth.
In the second stage, we perform self-supervised fine-tuning on data from the new domain.

Like MVSNet, our multi-view stereo reconstruction network compares image features in cost volumes.
This volume is refined with a set of 3D convolutions and we infer a preliminary depth map by neural regression from this refined volume. 
Different to the probability map for the depth as in MVSNet, we learn a confidence mask which is utilised to weight pixels for the self-supervised loss in order to compensate for outliers such as occlusions. 

We demonstrate our adaptive learning approach by training on the BlendedMVS~\citep{BlendedMVS} dataset which contains a large collection of outdoor scenes (e.g. views of buildings, architecture etc.) and indoor scenes. 
We fine-tune our pre-trained model using self-supervised training on the DTU dataset~\citep{DTU1} which consists of high resolution close scans of objects with different environment and lighting conditions. 
We evaluate our method on the DTU evaluation split and compare our approach to state-of-the-art MVS approaches and variants of our method such as fine-tuning without meta-learning. 
We demonstrate that meta-learning indeed helps to improve accuracy of MVS over naive fine-tuning.
Our approach improves reconstruction results over a self-supervised baseline method. 
In our experiments, it does even compare well with several previous supervised and classical methods in certain metrics.

In summary, our contributions are
\begin{itemize}[noitemsep,topsep=0pt]
	\item We propose a novel meta-learning scheme for adaptive learning of multi-view stereo reconstruction which improves self-supervised domain adaptation.
	\item We extend MVSNet to learn a confidence mask for per-pixel weighting for self-su\-pervised learning which handles outliers such as occlusions.
	\item We demonstrate that our meta-learning approach can improve self-supervised domain adaptation performance over naive pre-training in a supervised way. Our domain-adapted self-supervised multi-view stereo reconstruction achieves improved performance over a self-supervised MVS baseline.
\end{itemize}

\section{Related work}
{\bf Optimization-based Approaches.}
Multi view stereo estimation is one of the classical problems in computer vision with copious amount of research literature (see for example~\cite{survey1} for a survey).
State of-the-art systems such as COLMAP~\cite{colmap}, MVE~\cite{fuhrmann2014_mve} or PMVS~\cite{furukawa2} perform sparse structure from motion from collections of images to estimate sparse point cloud reconstruction, camera view poses and calibration parameters. Dense reconstruction is typically performed in a subsequent step, for instance, using patch-based surface representations and region-growing~\cite{furukawa2,grail,wei} or energy-minimization methods~\cite{pons2007,cremers2011}. 
Dense 3D surface reconstruction can be obtained by fusing depth maps in a 3D representation such as volumetric signed distance functions~\cite{voxfuse} or extracting meshes using point-cloud based surface reconstruction techniques~\cite{calakli2011_ssd,kazhdan2006_psr,ummenhofer2017_billion}.
A major problem of conventional multiview stereo approaches is that they are texture dependent and handcrafting good patch similarity measures is difficult. 

{\bf Supervised Learning Approaches.}
Early supervised deep learning methods learn similarity measures for patches from multiple views, for instance using Siamese network architectures~\cite{patchlearned,patch2learn}. 
More recent architectures~\citep{huang2018deepmvs,MVSNet,im2019dpsnet} integrate disparity plane sweeping directly into the deep neural network architecture and compare pixel locations based on learned deep feature representations. 
We also follow this approach with our architecture which is based on MVSNet~\cite{MVSNet}.
We extend the architecture with the prediction of a confidence mask and use it for meta-learning a model for domain adaptation using self-supervised training.
Recently, also methods have been proposed using recurrency~\cite{rmvsnet}, volumetric fusion~\cite{Paschalidou_2018_CVPR,Surfacenet}, or deep learning on point sets~\cite{PointMVSNet}.

{\bf Self-supervised Learning Approaches.}
One of the major problems of supervised techniques is the unavailibility of sufficiently large scale multi view stereo datasets with accurate depth map ground truth. In order to compensate for that, self-supervised multi-view stereo techniques have been developed very recently~\citep{khot,MVS2}. 
The basic idea behind these approaches is to use the predicted depth to synthesize stereo images or images in a temporal window and train the networks for photoconsistent estimates. The camera view poses are either known by different means or have to be estimated concurrently.
One of the major problems of both supervised and self-supervised learning approaches is that they typically do not generalize well to novel domains.

{\bf Meta-Learning.}
Recent developments in meta learning ~\cite{MAML} have demonstrated methods that efficiently adapt to novel tasks for supervised regression and reinforcement learning. 
The main idea behind model agnostic meta learning (MAML) is to train the model parameters in such a way that the network can better generalise to a new task through fine-tuning. 
Previous work on adaptive learning of stereo disparity estimation~\cite{L2A} has utilised this meta-learning and have shown how feature representations can be learned for self-supervised learning and improved generalization on new datasets. 
We propose to learn adaptive feature representations for self-supervised multi-view stereo reconstruction through meta-learning.
We develop extensions to a network architecture based on MVSNet~\cite{MVSNet} with which the model learns to mask uncertain predictions due to outliers such as occlusions.
This assists the self-supervised fine-tuning on new domain data.

\section{Methodology}
Meta learning aims at training a learning architecture for fast adaptability to new tasks. 
To this end, the model is trained on a set of different tasks during the meta learning phase. 
In our context, tasks correspond to self-supervised learning in different environments and conditions (i.e. domains).

The methodology can be summarized in two stages. The model is trained on a larger dataset with ground-truth depth in the first
stage using meta-learning. 
In our experiments, we use the BlendedMVS dataset~\citep{BlendedMVS} which consists of indoor and outdoor scenes with varying environment
conditions - making it ideal for domain adaptation. 
The model is trained on the training split by first
updating cloned model parameters using the self-supervised photometric losses for \textit{k} 'tasks', where a task refers to multi-view reconstruction of one of the k different scenes. The actual model parameters are then in turn updated using the supervised loss in Eq.~\eqref{eq:2} on the validation split, which involves the cloned and updated parameters from the previous step. The model trains network parameters to adapt well by self-supervised training. This is guided through the outer-loop supervised training (see Alg.~\ref{alg:meta_learning_algorithm}).
The second stage involves fine-tuning the model obtained in the first stage using self-supervised learning on the training data of the target domain dataset (DTU~\citep{DTU1} in our experiments). 
We provide detailed explanation of the methodology in the following subsections.

\subsection{Meta Learning for Self-supervised Multi-View Stereo}

\begin{figure}
	\centering     
	
	\includegraphics[width=1\textwidth]{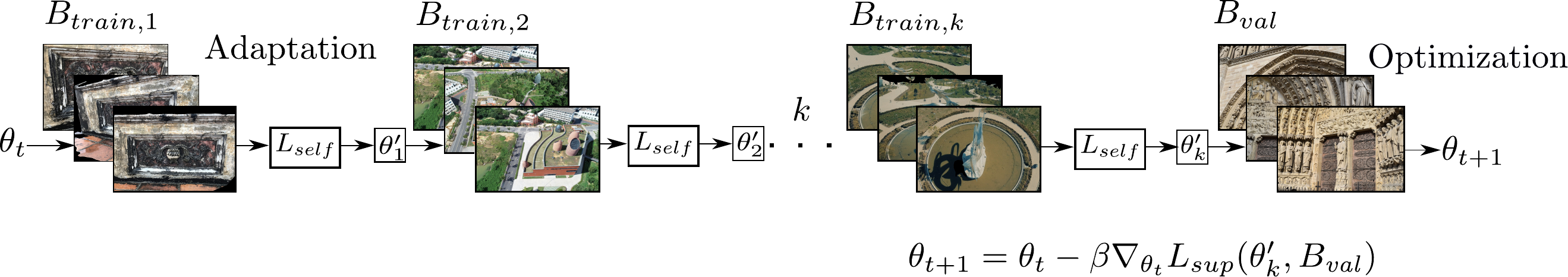}
	\caption{Meta-learning for self-supervised multi-view stereo. During a meta-learning iteration, adaptation is performed on $k$ multi-view stereo reconstruction tasks with a self-supervised loss ($L_{self}$). The adapted parameters $\theta'_{k}$ are evaluated and the base model parameters $\theta_{t}$ are optimized on a validation set using a supervised loss $L_{sup}$ to learn a better starting point for self-supervised parameter adaptation. For the new domain, the resulting base model trained through meta-learning is fine-tuned with self-supervised training.}

	\label{fig:meta training}
\end{figure}

Our meta-learning algorithm for self-supervised multi-view stereo is summarized in Alg.~\ref{alg:meta_learning_algorithm} and illustrated in Fig.~\ref{fig:meta training}.
We split the training dataset $D$ into a training and a validation split $D_{\mathit{train}}$ and $D_{\mathit{val}}$, the latter with $m$ multi-view examples.
Each example consists of a reference view (image with camera pose) and $N$ neighbouring views.

We adapt the base model parameters $\theta$ for $k$ multi-view examples $B_{{\mathit{train}},i} \subset D_{\mathit{train}}$ each consisting of one reference view and $N$ neighbouring views of a scene using a self-supervised loss ($L_{self}$, eq.~\eqref{eq:5}).
Starting from the base parameters $\theta'_0 = \theta$, for each multi-view example $i$ we perform the gradient update steps
\begin{equation}
\label{eq:1}
	\theta'_i = \theta'_{i-1} - \alpha \nabla_{\theta'_{i-1}} L_{self}(\theta'_{i-1}, B_{\mathit{train},i}),
\end{equation}
where $\alpha$ is a learning rate.

The base model parameters are optimized to improve the quality of the updated model parameters $\theta'_k$ with a supervised loss $L_{sup}$ (eq.~\eqref{eq:9}) on a sampled multi-view example $B_{\mathit{val}} \subset D_{\mathit{val}}$ consisting of one reference view and $N$ neighbouring views from the validation split,
\begin{equation}
\label{eq:2}
	min_{\theta}( L_{sup}( \theta'_k, B_{\mathit{val}} ) ).
\end{equation}
Note that $\theta'_k$ is a function of $\theta$ through the updates in Eq.~\eqref{eq:1}.
The supervised loss measures the discrepancy between the predicted and the ground-truth depth. 

The intuition behind this two-step update scheme is that the base model parameters are changed to a better starting point for learning model parameters on different domains with the self-supervised loss.
For a new dataset, we use the base parameters $\theta$ for fine-tuning to the new domain (i.e an entirely unseen dataset with different conditions and environment) using self-supervised training. 

\begin{algorithm}
 \KwData{Dataset split $D_{train}$, $D_{val}$, hyperparameters $k$, $\alpha$, $\beta$ }
 Initialize base model parameters $\theta$ \;
 \While{not converged}{
	Sample $k$ multi-view examples $B_{\mathit{train},i} \subset D_{\mathit{train}}$ \;
	Initialize model parameters $\theta_0 = \theta$ \;
	\For{$i \in \left[1,\ldots,k\right]$}{
		Compute adapted model parameters $\theta'_i = \theta'_{i-1} - \alpha \nabla_{\theta'_{i-1}} L_{self}(\theta'_{i-1}, B_{\mathit{train},i})$ \tcp*[l]{Adaptation}
	}
  Sample batch $B_{\mathit{val}} \subset D_{\mathit{val}}$ \;
  Perform gradient descent step on base model parameters $\theta$ to minimize $L_{sup}( \theta'_k, B_{\mathit{val}} )$ with learning rate $\beta$ \tcp*[l]{Optimization}
	}
 \label{alg:meta_learning_algorithm}
 \caption{Adaptive learning for self-supervised multi-view stereo.}
\end{algorithm}

\subsection{Network Architecture}
While our adaptation module is model-agnostic, we base our network architecture on the MVSNet~\cite{MVSNet} model.
MVSNet has demonstrated state-of-the-art performance for both supervised and self-supervised~\cite{khot} training. 
Besides changing the training schemes with our meta-learning approach, we also augment the network with predicting confidence masks which are in turn used for self-supervised fine-tuning on novel domains (additional details can be found in the supplementary material). 
For details on the base network, readers are encouraged to refer to MVSNet~\cite{MVSNet}. 


\subsection{Learning Confidence Masks for Self-supervised Domain Adaptation}
A major problem in learning multi-view stereo is to handle (dis-)occlusions and out-of-image projections correctly when quantifying the loss on the predicted depth maps.
We take inspiration from~\cite{L2A} to learn a confidence mask during meta-learning which is used to improve the fine-tuning of the network on the new domain.
While the approach in~\cite{L2A} has been proposed for learning dense reconstruction from stereo images of a constant-baseline stereo rig, multi-view stereo poses additional challenges due to the varying baselines between reference image and neighbouring frames.

\begin{figure}
	\centering     
	
	\includegraphics[width=1\textwidth]{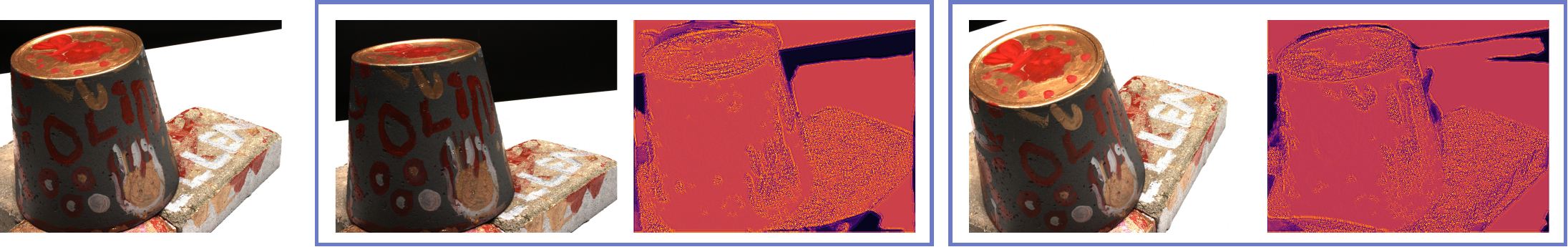}
		\caption{\textit{From left to right}: reference image frame, first neighbouring frame, predicted confidence mask for first frame, second neighbouring frame and predicted confidence mask for second view. Different pairs of reference and neighbouring frames have different outliers such as occlusions, reflections, etc. We learn a confidence mask during meta-learning to downweight uncertain pixels for the self-supervised training and fine-tuning on new domains (darker color correspond to lower confidence in the visualization).}
	\label{fig:mask}
	\label{fig:confmask}
\end{figure}

Our network learns a confidence mask for each pair of reference image and neighboring frame in a multi-view training set.
Fig.~\ref{fig:confmask} provides an example of the confidence masks learned by our approach for different neighbouring views. 
The out-of-image projection mask $C_{\mathit{proj}}: \Omega \rightarrow \left[ 0, 1 \right]$ can be directly determined from the relative camera pose between the views and the predicted depth map.
We train an additional component of our network architecture during meta-learning which predicts a confidence mask $C_{\tau}: \Omega \rightarrow \left[ 0, 1 \right]$ with learnable parameters $\tau$ for learning to downweight pixels in the loss at occlusions and other outliers.
The final per-pixel mask is obtained by the product of the two masks at each pixel.
The masks are used for the self-supervised loss to compensate for occlusions due to view pose changes. 
Note that the parameters $\tau$ are included into $\theta$ and updated during the meta-learning stage. 
They are held fixed when fine-tuning on a new domain dataset in the second stage.

The confidence mask network is a 4-layer CNN with sigmoid activation at the end to generate values between 0 and 1. The photometric warping error between reference $I_{ref}$ and neighboring image $I^{i}$, and the out-of-image projection mask are concatenated and used as an input to the network that predicts the confidence mask. 
Please refer to the supplementary material for further details on the confidence mask subnetwork details and how it is integrated in our network architecture. 



%

\subsection{Training Losses}
{\bf Self-supervised Losses.} 
Self-supervised losses are used for adaptation during meta-learning and for fine-tuning on the new domain. 
The self-supervised loss comprises two components,
 \begin{equation}
 \label{eq:5}
L_{self}( \theta, B ) = L_{\text{recon}}( \theta, B ) + \gamma_{\text{smooth}} L_{\text{smooth}}( \theta, B ),
\end{equation}
a reconstruction loss $L_{\text{recon}}$ and a smoothness loss $L_{\text{smooth}}$, where $\theta$ are network parameters and $B$ is a data example consisting of a reference frame and $N$ neighbouring frames.

The reconstruction loss measures the image-based consistency between the reference and the $N$ neighbouring views given their relative camera pose and the predicted depth map,
\begin{multline}
\label{eq:6}
L_{\text{recon}}( \theta, B ) = \sum_{i=1}^{N} \gamma_{\text{photo}} \left\|C_{\tau}^i( \theta, B ) \odot C_{\mathit{proj}}^{i} \odot \left( I_{ref} - I_{warped}^{i}( \theta, B ) \right) \right\|_1  +\\ \gamma_{\text{ssim}} \left\| 1 - SSIM( C_{\mathit{proj}}^{i} \odot I_{ref} , C_{\mathit{proj}}^{i} \odot I_{warped}^{i}( \theta, B ) ) \right\|_1,
\end{multline}
where $\gamma_{\text{photo}}$ and $\gamma_{\text{ssim}}$ are weighting factors and $\odot$ denotes pixel-wise multiplication.
We use a combination of a photoconsistency measure and the structural similarity index (SSIM~\cite{ssim}).
The reference image is $I_{ref}$, $I_{warped}^{i}( \theta, B )$ is the $i^{th}$ neighbouring frame warped to the reference frame given the predicted depth by the network and known camera parameters. $C_{\mathit{proj}}^{i}$ is the out-of-image projection mask which excludes the out of bound pixels while warping and $C_{\tau}^{i}( \theta, B )$ is the predicted confidence mask for the $i^{th}$ frame.
The structural similarity index~\cite{ssim} quantifies the similarity between $I_{ref}$ and $I_{warped}$ in patches centered at the pixels, and has been used in the literature~\citep{khot,monodepth1} since it measures texture similarity while being more robust to lighting changes than the photometric L1-loss. 

An edge-dependent smoothness prior on the predicted depth maps with respect to the reference image is applied in order to encourage smoothness of the depth map. The smoothness loss for the predicted depth map $D( \theta, B )$ is
\begin{equation}
  \label{eq:8}
  L_{\text{smooth}}( \theta, B ) = \sum_{(x,y)} \left | \partial_x D_{x,y}( \theta, B ) \right | e^{-\left \|
        \partial_x I_{x,y} \right \|_2} + \left |
        \partial_y D_{x,y}( \theta, B ) \right | e^{-\left \| \partial_y I_{x,y} \right \|_2},
\end{equation}        
where $x,y$ range over the pixels in the reference frame.

{\bf Supervised Loss.} For evaluation during meta-training, we use an L1 supervised loss on the depth map $D( \theta, B )$ predicted by the network to compare it with the ground truth $D_{\text{gt}}$, 
\begin{equation}
 \label{eq:9}
	L_{sup}( \theta, B ) = \left\| D( \theta, B ) - D_{gt} \right\|_1 .
\end{equation}

\section{Experiments}

We evaluate our approach on a large-scale MVS dataset with ground-truth for the meta-learning stage and demonstrate domain adaptation on a smaller-scale MVS dataset from a different domain.
For meta-learning, we use the BlendedMVS dataset~\citep{BlendedMVS} which has a mix of outdoor and indoor scenes.  
The dataset contains over 17k
high-resolution images covering a variety of scenes, including cities, architectures, sculptures and small objects. 
The dataset is divided into training and validation sets which we use for the meta-learning. 
Domain adaptation is tested on the DTU~\citep{DTU1} dataset, where we fine-tune the model on the training split and evaluate its final performance on the test split.
The DTU scans consist of different objects in a different indoor environment with varied lighting conditions. 

\subsection{Training Details}
The number of neighbouring frames $N$ is 2 for meta-learning and fine-tuning. 
The model is tested with $N=4$ frames. 
The number of depths ($d = 256$), input resolution ($H=512,W=640$) and output depth resolution ($H=128,W=160$) are initialized as in the original MVSNet setup for fair comparison~\citep{MVSNet}. 
Learning rates are selected as $\alpha = 10^{-4}$ and $\beta = 10^{-4}$. 
The model is fine-tuned with a learning rate of $10^{-7}$ and a batch size of 4 multi-view examples with one reference frame and $N$ neighbouring frames each.
The self-supervised loss weights are set to $\gamma_{\text{photo}}=5$,$\gamma_{\text{ssim}}=1$ and $\gamma_{\text{smooth}}=0.01$. 
For meta-learning we use $k=3$ multi-view examples in each update cycle. 
The meta-training and testing have been performed on the same hardware configuration (4 NVidia Titan RTX GPUs) using a PyTorch implementation. 
We used the Learnable~\citep{learn2learn2019} library for implementing first-order MAML. 

\subsection{Depth Map Fusion}
Similar to MVSNet~\cite{MVSNet}, we fuse the predicted depth maps into point cloud reconstructions using~\cite{merrell2007_visbasedfusion}\footnote{We use the open-source implementation at \url{https://github.com/xy-guo/MVSNet_pytorch} with its default parameter setting}. The method determines a subset of the images using the view selection score of COLMAP~\citep{colmap}. Their depth maps are projected to 3D points in a common coordinate frame. Matches of points in neighboring views are found through reprojection into the images. Points with reprojection distance error with threshold $<1$ and relative depth difference with threshold $<0.01$ are averaged to obtain the final point cloud.
We reconstruct point clouds by fusing the generated depth maps for those pixels with confidence above threshold $>0.8$.

\subsection{Quantitative Results}
The fine-tuned model is evaluated on the DTU test split~\citep{MVSNet,Surfacenet}. 
We use the evaluation metrics as in~\citep{DTU2}.
The \textit{accuracy} distance metric is measured as the distance from the
estimated reconstruction to the ground-truth, encapsulating the accuracy of the estimated points. The \textit{completeness} is measured as the distance from
the ground-truth reconstruction points to the estimated reconstruction, encapsulating how much of the surface is captured by the
MVS reconstruction.
\textit{Overall} is the mean of \textit{accuracy} and \textit{completion} (see Table \ref{tab:results1}).
Additionally, we report the \textit{overall F-score} metric~\citep{Tanks} at inlier thresholds of 1\,mm and 2\,mm. 
We utilize~\citep{Open3d} for calculating the precision and recall (see Table \ref{tab:results1}: \%-age (percentage) columns ).
The F-score is the harmonic mean of precision and recall.

The results in Table~\ref{tab:results1} demonstrate that our method can improve results over its self-supervised baseline MVSNet in~\citep{khot}. 
It is second to~\citep{MVS2} in terms of overall metric among self-supervised methods. Filtering
with the confidence mask can lead to higher accuracy in favor of lower completeness. 
Note that our method
achieves state-of-the-art results in the overall F-score measures at 1\,mm and 2\,mm inlier threshold compared to self-supervised and classical methods. 
Remarkably it fares similar to one of the supervised methods (SurfaceNet) in several metrics. 

\begin{table}[tb]
\resizebox{\columnwidth}{!}{
\begin{tabular}{lccccccccc}
\toprule
 method & acc. & comp. & over. & prec. & rec. & over. F& prec. & rec. & over. F\\ 
 \cmidrule(lr){2-4}  \cmidrule(lr){5-7}  \cmidrule(lr){8-10}
  &  &   &  &   & (1 mm) in \% &  &  & (2mm) in \% &\\\midrule
  
Camp~\cite{Camp} (C)        	 & 0.835 & 0.554 & 0.695 & 71.75 & 64.94 & 66.31  & 84.93 & 69.93 & 74.36          \\
Furu~\cite{Furu}(C)            	 & 0.612 & 0.939 & 0.775 & 69.55 & 61.52 & 63.26  & 77.3 & 64.06 & 70.06          \\
Tola~\cite{Tola}(C)      		 & \textcolor{Blue}{0.343} & 1.19 & 0.766 &  \textcolor{Blue}{90.49} & 57.83 & 68.07  &  \textcolor{Blue}{92.35} & 60.01 & 72.75\\ \hline

MVSNet~\cite{MVSNet}(Sup DTU)          & 0.396 & 0.527 & 0.462 &86.46 &  71.13 &  75.69  & 91.06 &  75.70 &  80.25          \\

Ours (Sup PT bMVS, Sup FT DTU)          & 0.441 & \textcolor{Blue}{0.387} & \textcolor{Blue}{0.414} &83.55 &  \textcolor{Blue}{74.25} &  \textcolor{Blue}{76.93}  & 88.56 &  \textcolor{Blue}{77.63} &  \textcolor{Blue}{81.09}          \\  

Surfacenet~\cite{Surfacenet}(Sup DTU)  & 0.450 & 1.043 & 0.746 & 83.8 & 63.38 & 69.95  & 87.44 & 67.87 & 74.81          \\
\midrule
MVSNet~\cite{khot} (Self DTU)          & 0.881 & 1.073 & 0.977  & 61.54 & 44.98   & 51.98   & 85.15 & 61.08 & 71.13          \\
MVS2~\cite{MVS2}(Self DTU)              & 0.760 & \bf 0.515  & \textbf{0.633}& 70.56 & \textbf{66.12} & 68.27  & - & - & -  \\ \midrule
Ours (Meta PT bMVS, Self FT DTU)                    & \textbf{0.5942} & 0.7787 & 0.6865 & \textbf{80.18} & 63.58 & \textbf{68.67} & \textbf{90.95} & \textbf{69.08} & \textbf{76.22}\\ \bottomrule
\end{tabular}
}
\caption{\label{tab:results1} Evaluation scores for reconstruction metrics (C: classical, Sup: supervised, Self: self-supervised, Meta: meta-learning). PT: pre-trained, FT: fine-tuned. bMVS: trained on Blended MVS. DTU: trained on DTU. Lower score is better for accuracy (acc.), completeness (comp.) and overall (over.) metrics. Higher  score is better for precision (prec.), recall (rec.) and overall F-score (over. F) metric. Blue indicates best among all methods. Best results among methods trained self-supervised on DTU are shown in bold. Our approach demonstrates improved results over its self-supervised baseline MVSNet~\cite{khot}. Our method
achieves state-of-the-art results in the overall F-score measures at 1\,mm and 2\,mm inlier threshold compared to self-supervised and classical methods. We even fare similar to a supervised approach (SurfaceNet) in several metrics.  }
\end{table}

\begin{figure}[tb]
	\resizebox{\columnwidth}{!}{
		\begin{tabular}{ccccc}
			Ground Truth & MVSNet (Sup) & SurfaceNet (Sup) & Ours (Meta, Self) \\
			\includegraphics[trim={17cm 8cm 17cm 3cm},clip,width=.23\textwidth,height = 0.17\textwidth]{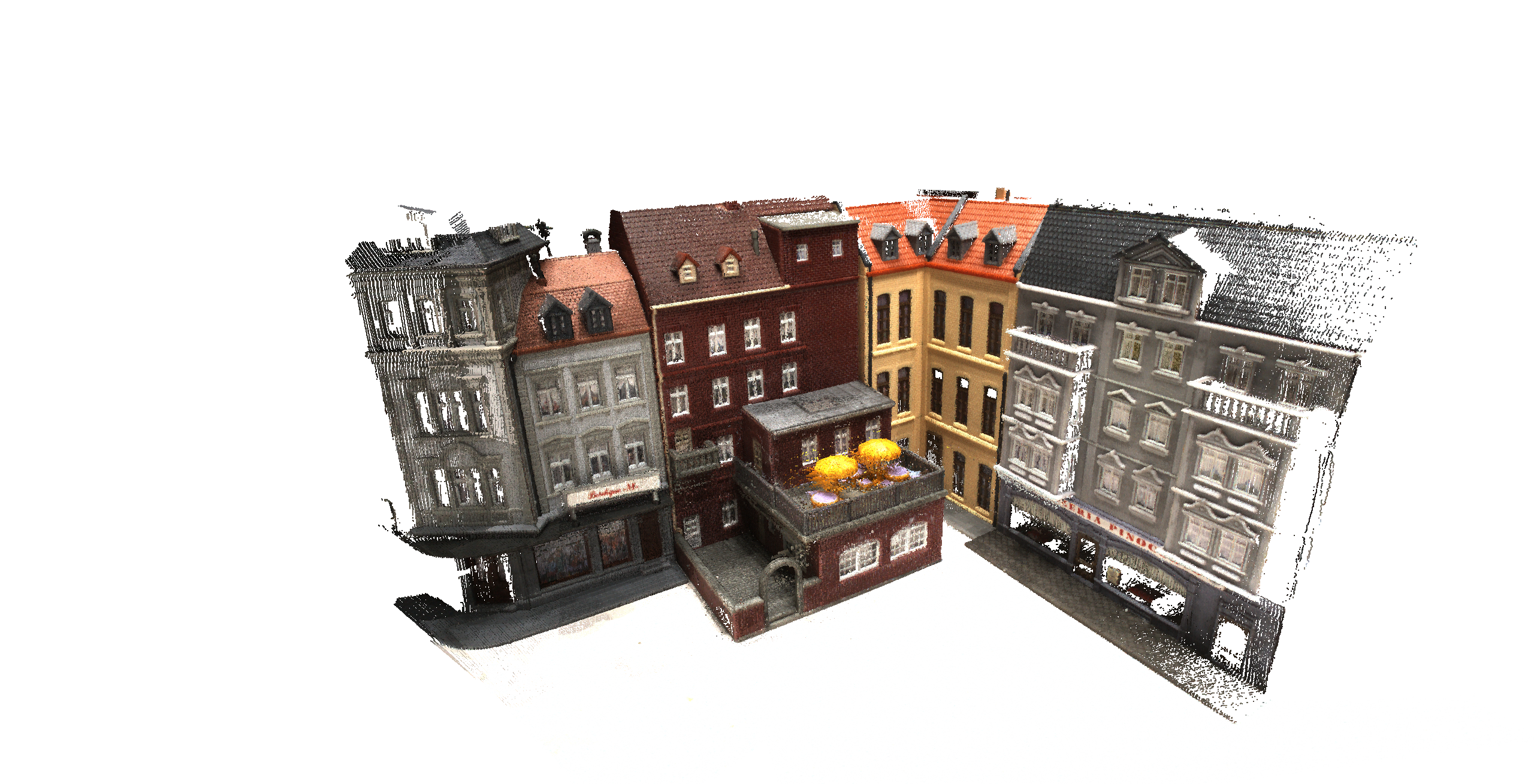} & 
			\includegraphics[trim={17cm 8cm 17cm 3cm},clip,width=.23\textwidth,height = 0.17\textwidth]{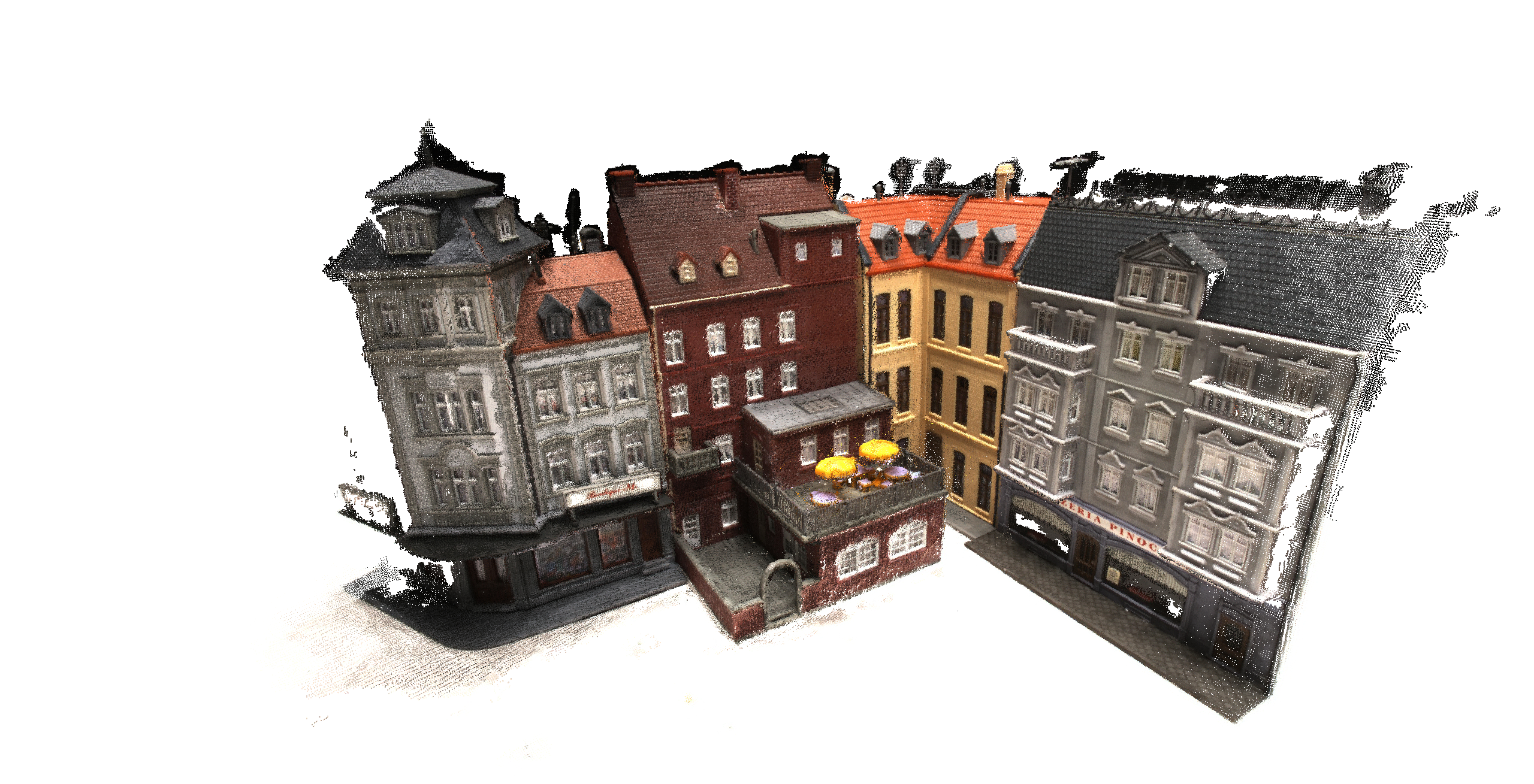} & 
			\includegraphics[trim={17cm 8cm 17cm 3cm},clip,width=.23\textwidth,height = 0.17\textwidth]{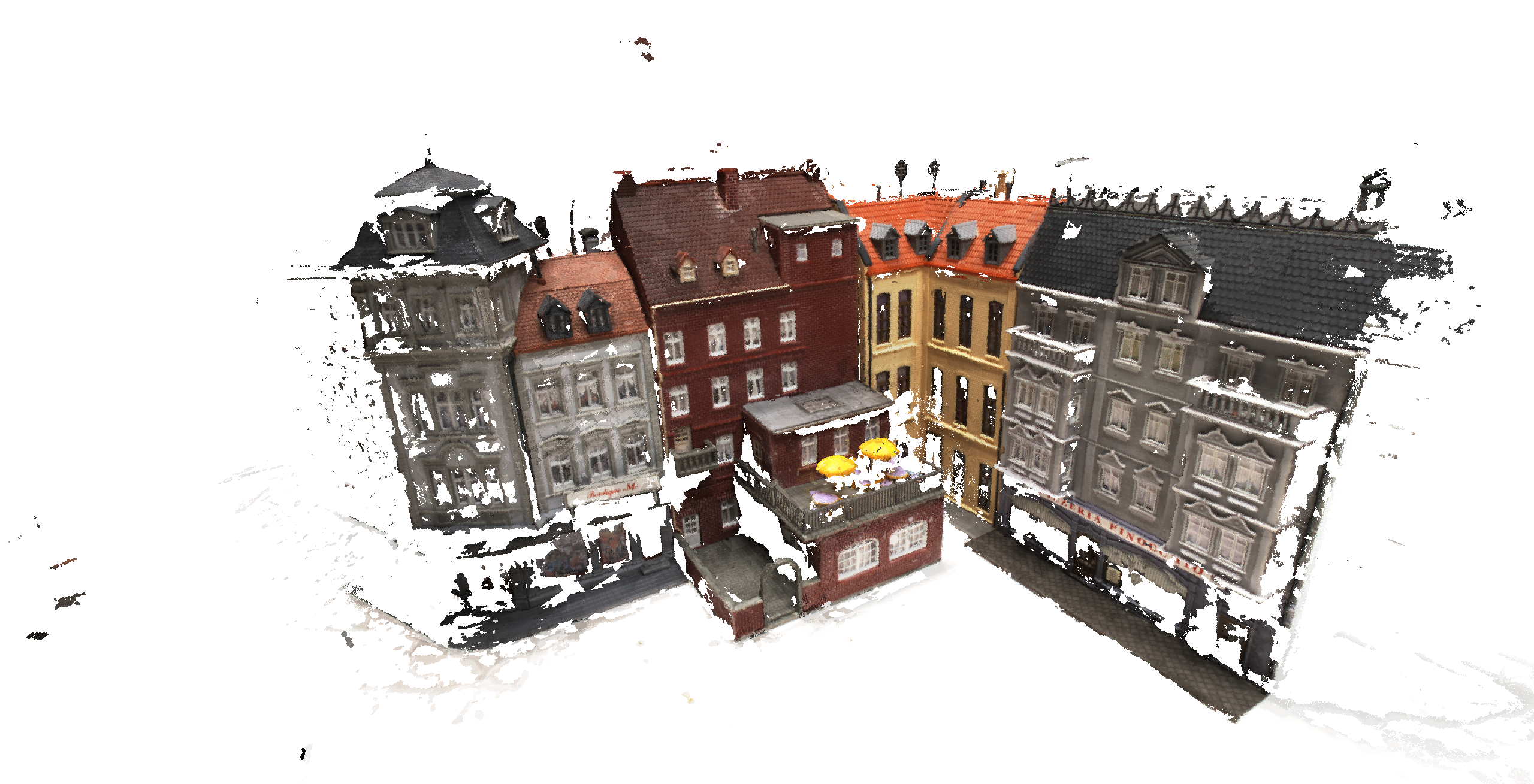} & 
			\includegraphics[trim={17cm 8cm 17cm 3cm},clip,width=.23\textwidth,height = 0.17\textwidth]{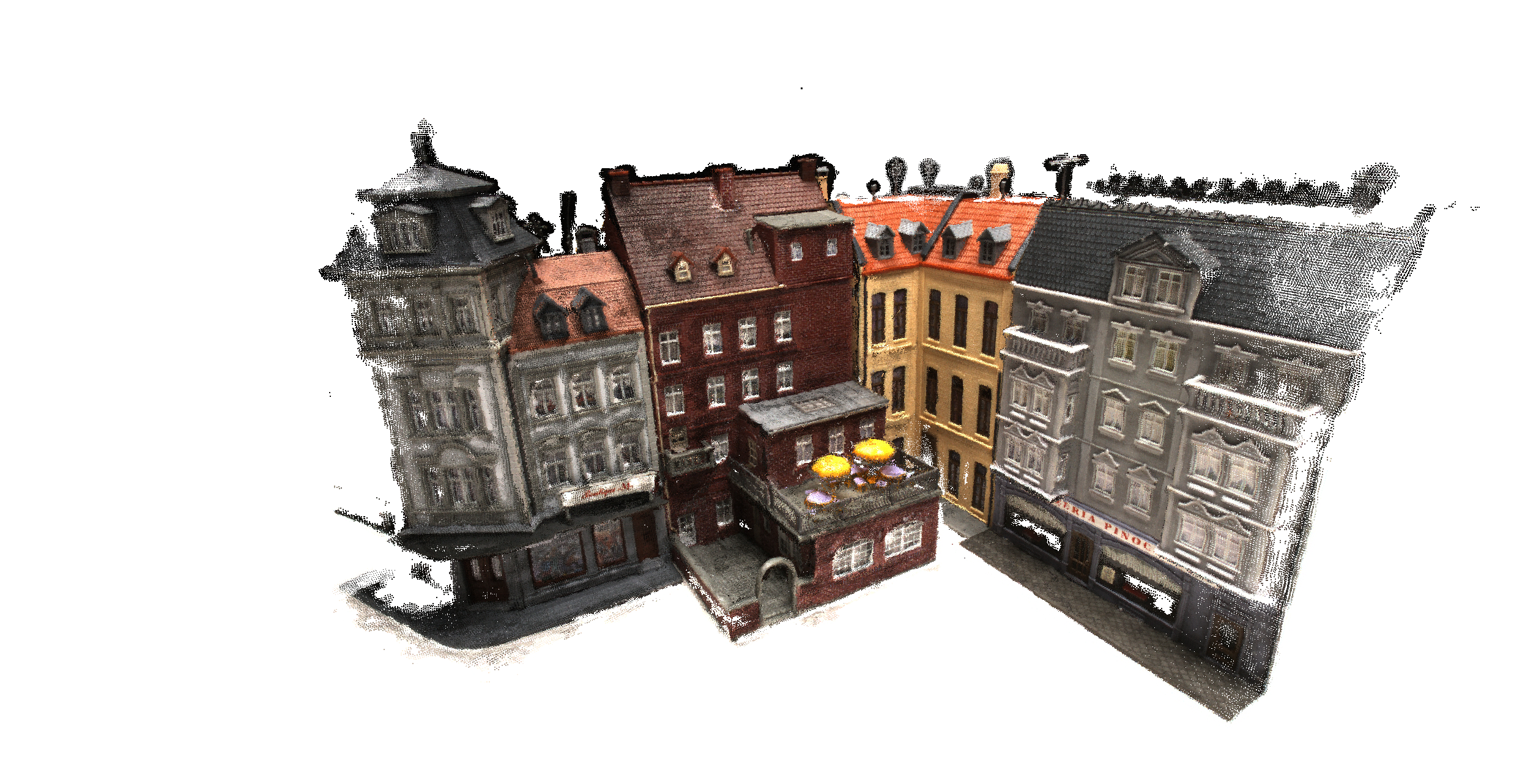} \\
			
			\includegraphics[trim={17cm 4cm 17cm 6cm},clip,width=.23\textwidth,height = 0.17\textwidth]{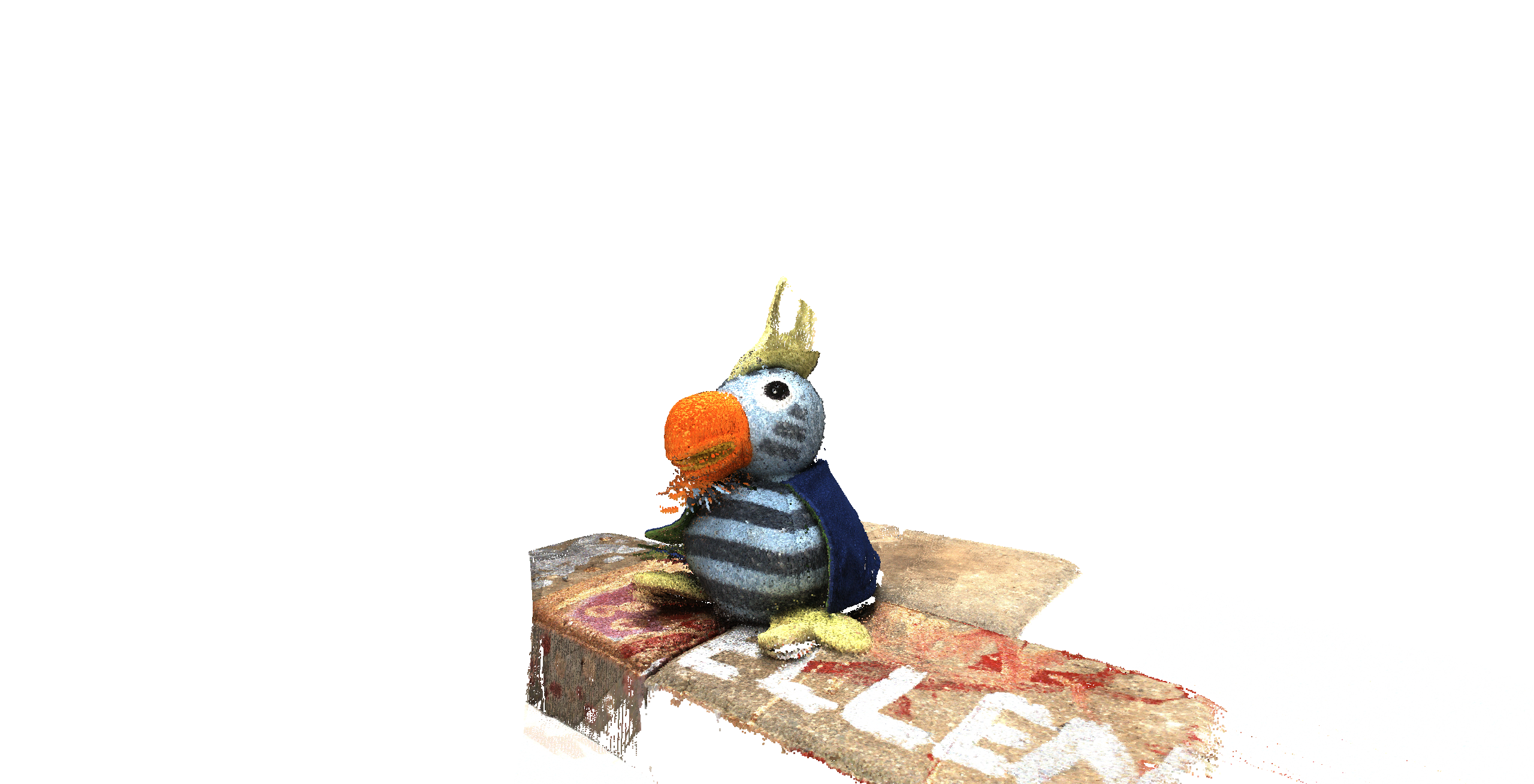} & 
			\includegraphics[trim={17cm 4cm 17cm 6cm},clip,width=.23\textwidth,height = 0.17\textwidth]{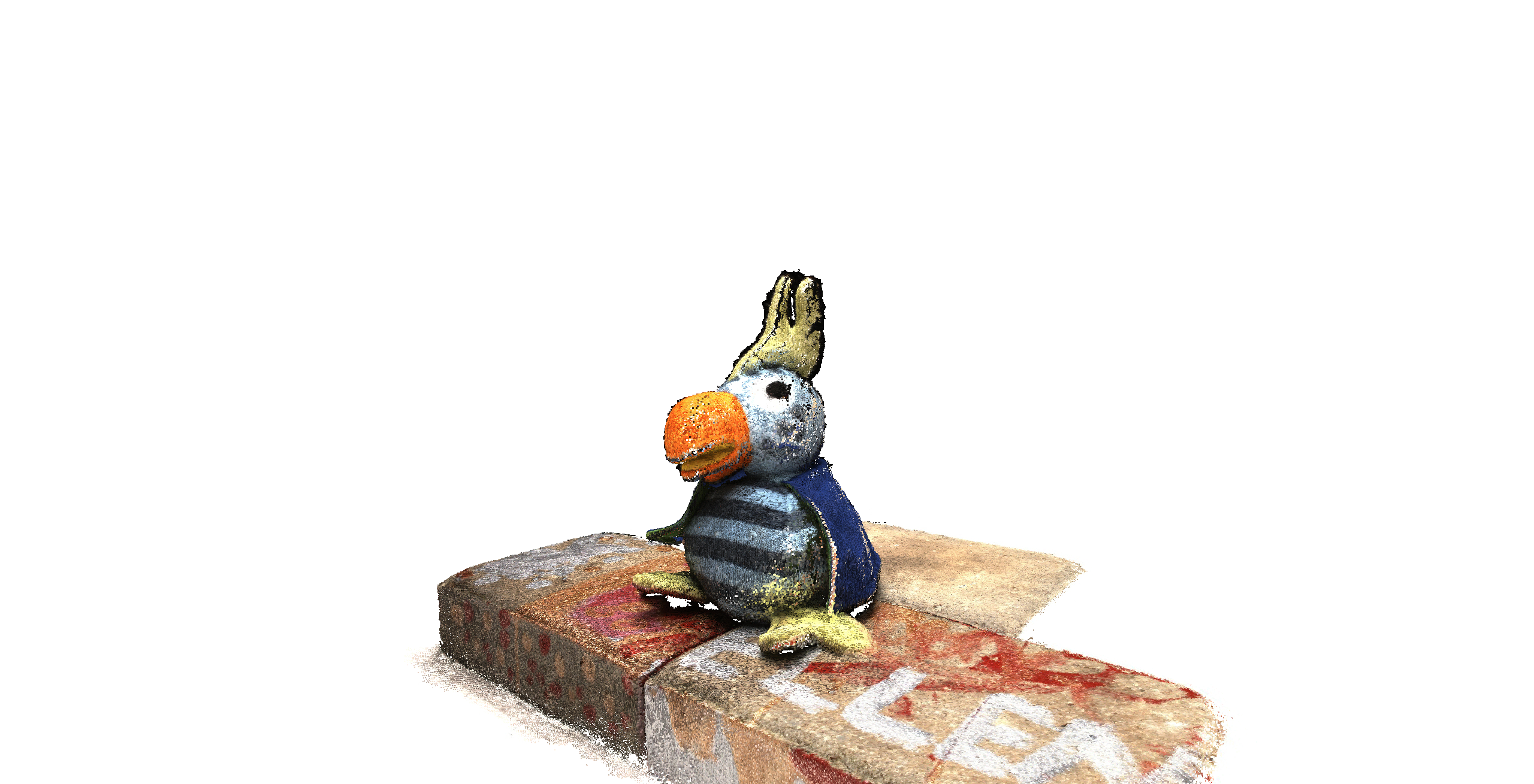} & 
			\includegraphics[trim={17cm 4cm 17cm 6cm},clip,width=.23\textwidth,height = 0.17\textwidth]{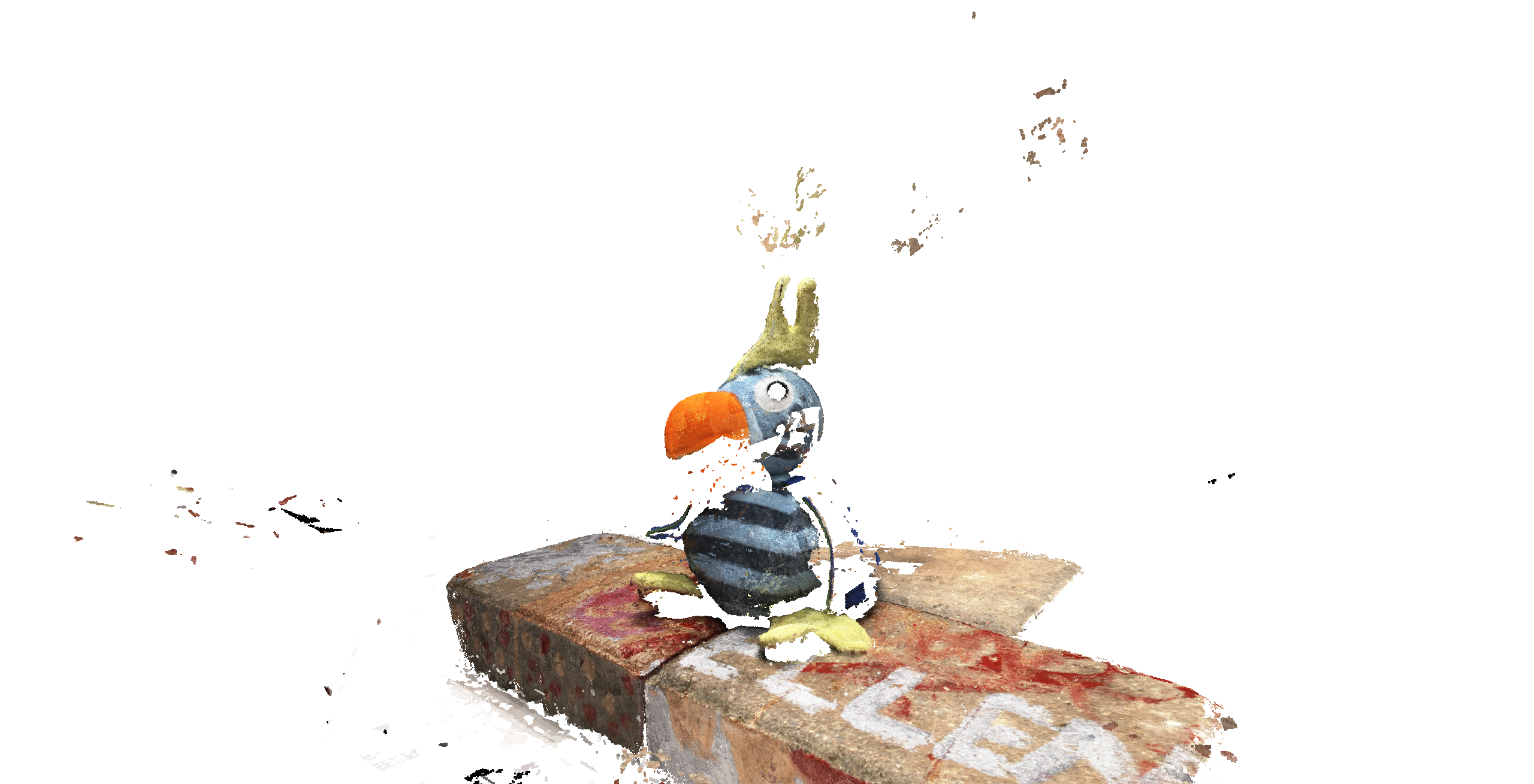} & 
			\includegraphics[trim={17cm 4cm 17cm 6cm},clip,width=.23\textwidth,height = 0.17\textwidth]{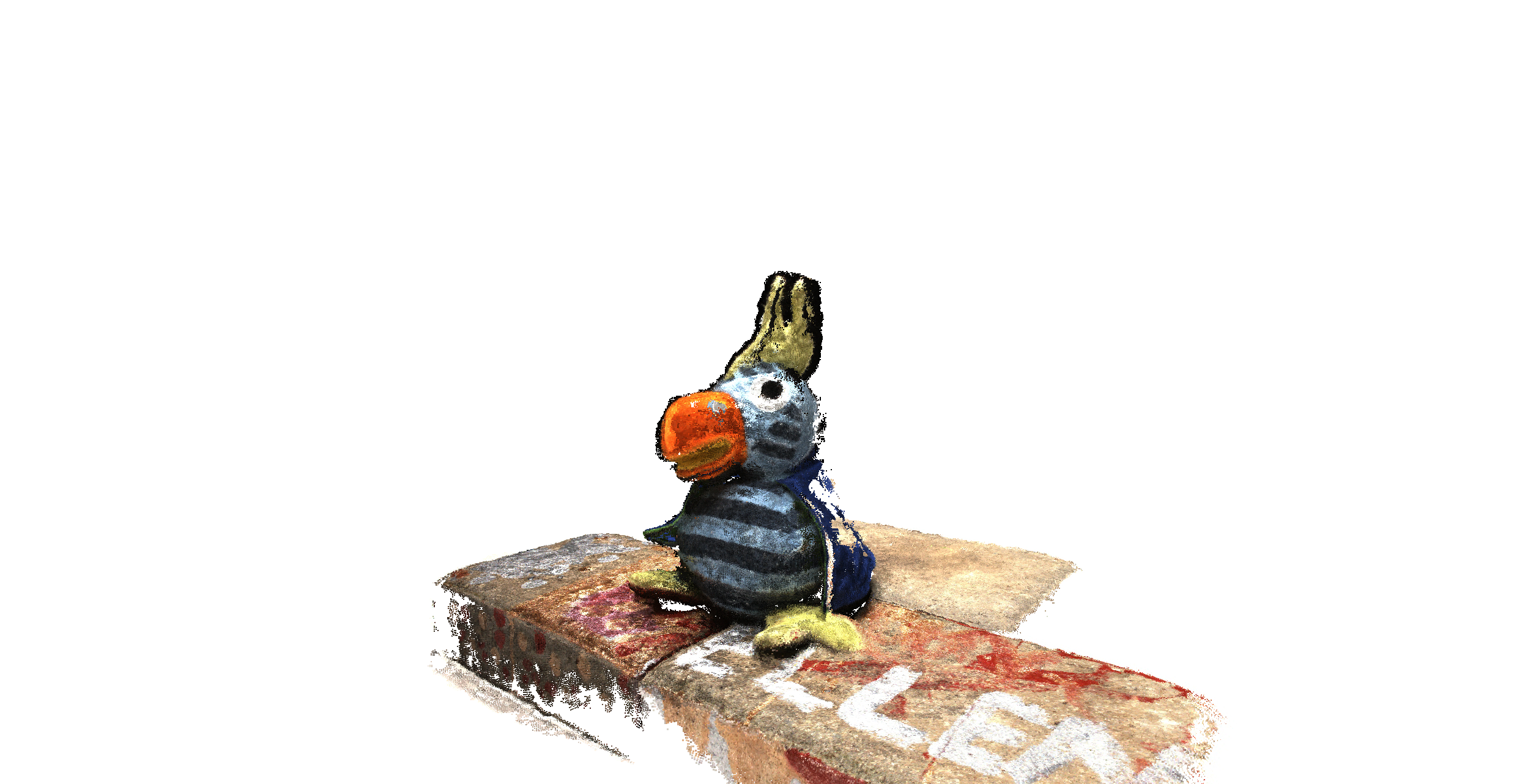} \\
			
			\includegraphics[trim={18cm 8cm 21cm 5cm},clip,width=.23\textwidth,height = 0.17\textwidth]{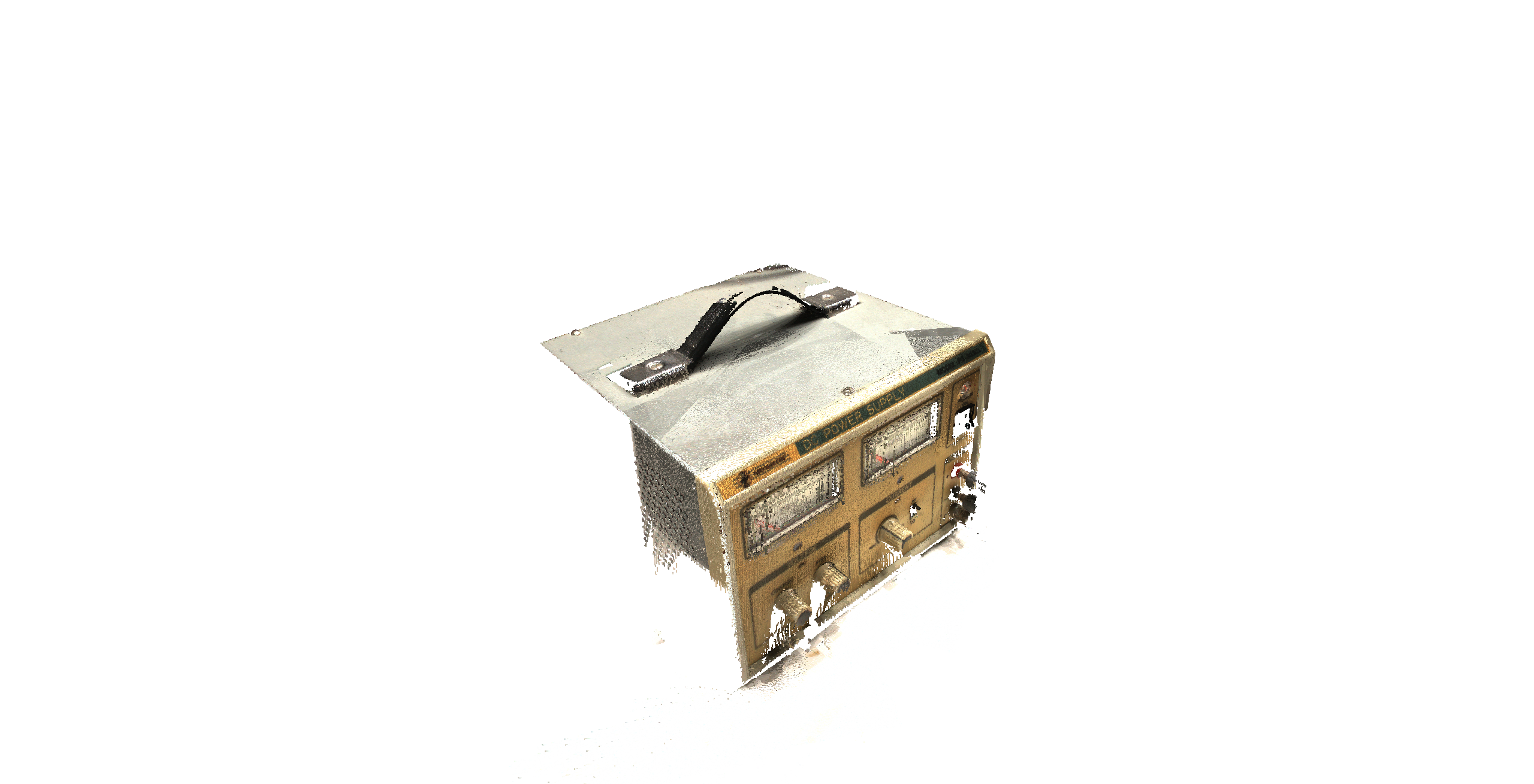} & 
			\includegraphics[trim={18cm 8cm 21cm 5cm},clip,width=.23\textwidth,height = 0.17\textwidth]{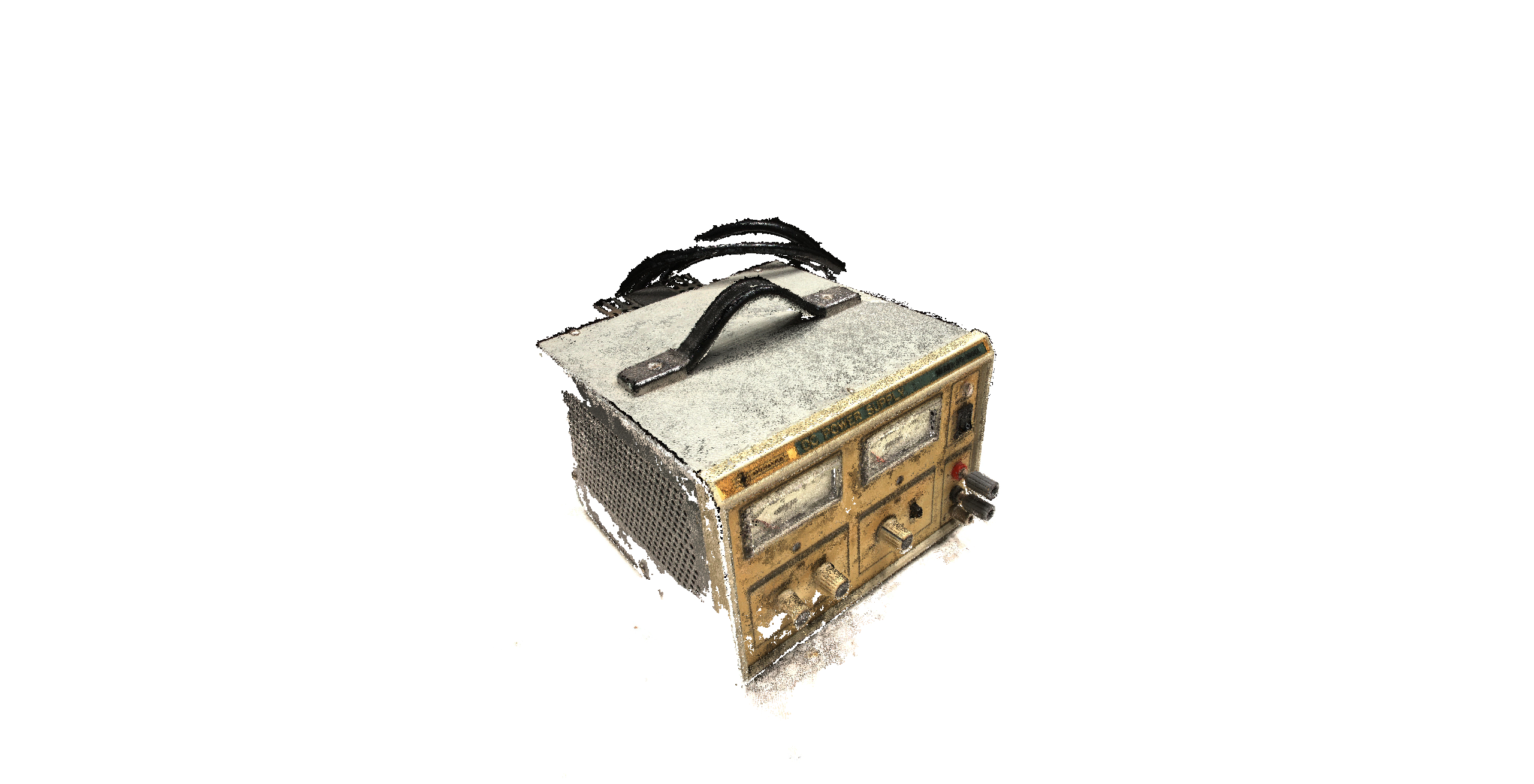} & 
			\includegraphics[trim={18cm 8cm 21cm 5cm},clip,width=.23\textwidth,height = 0.17\textwidth]{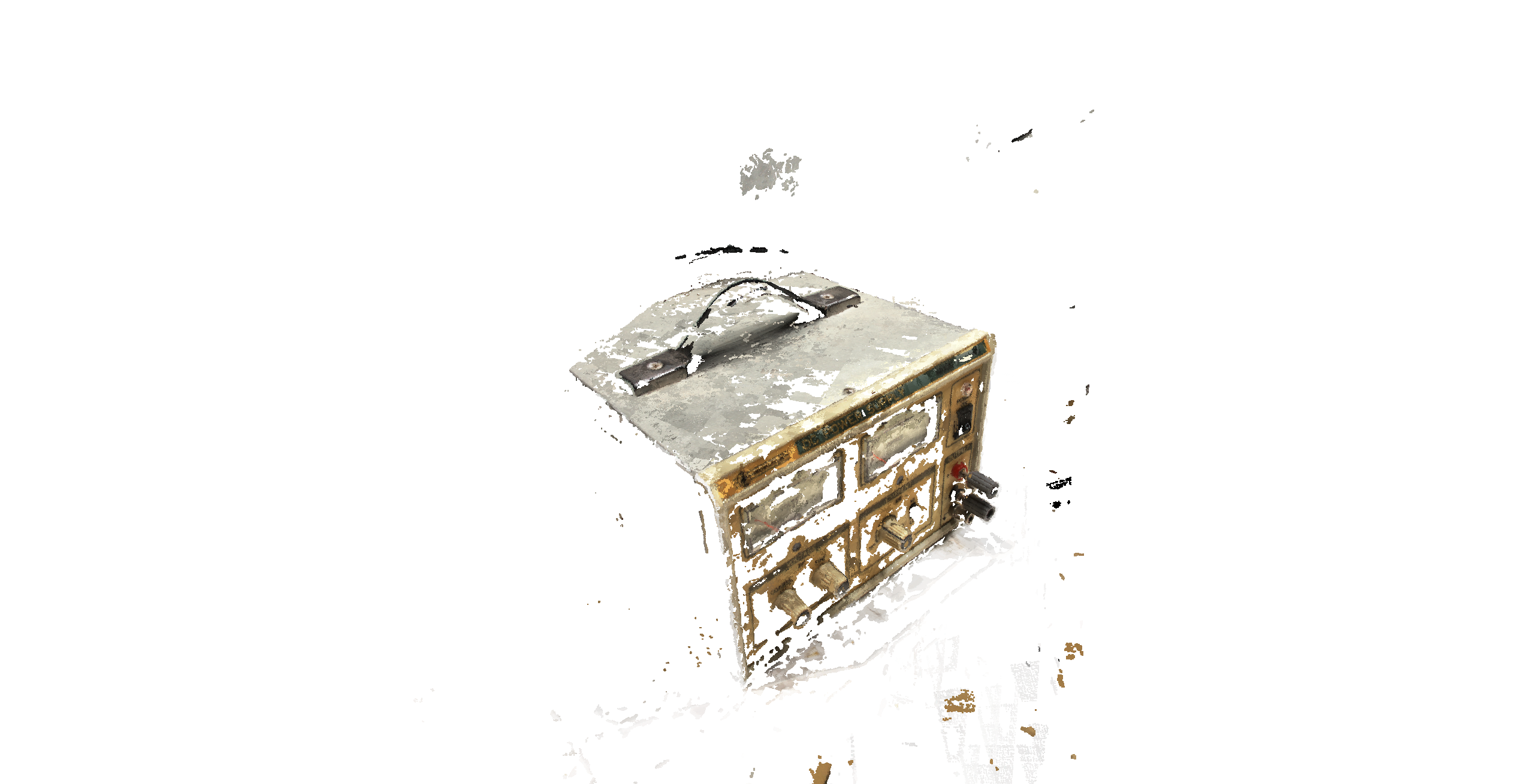} & 
			\includegraphics[trim={18cm 8cm 21cm 5cm},clip,width=.23\textwidth,height = 0.17\textwidth]{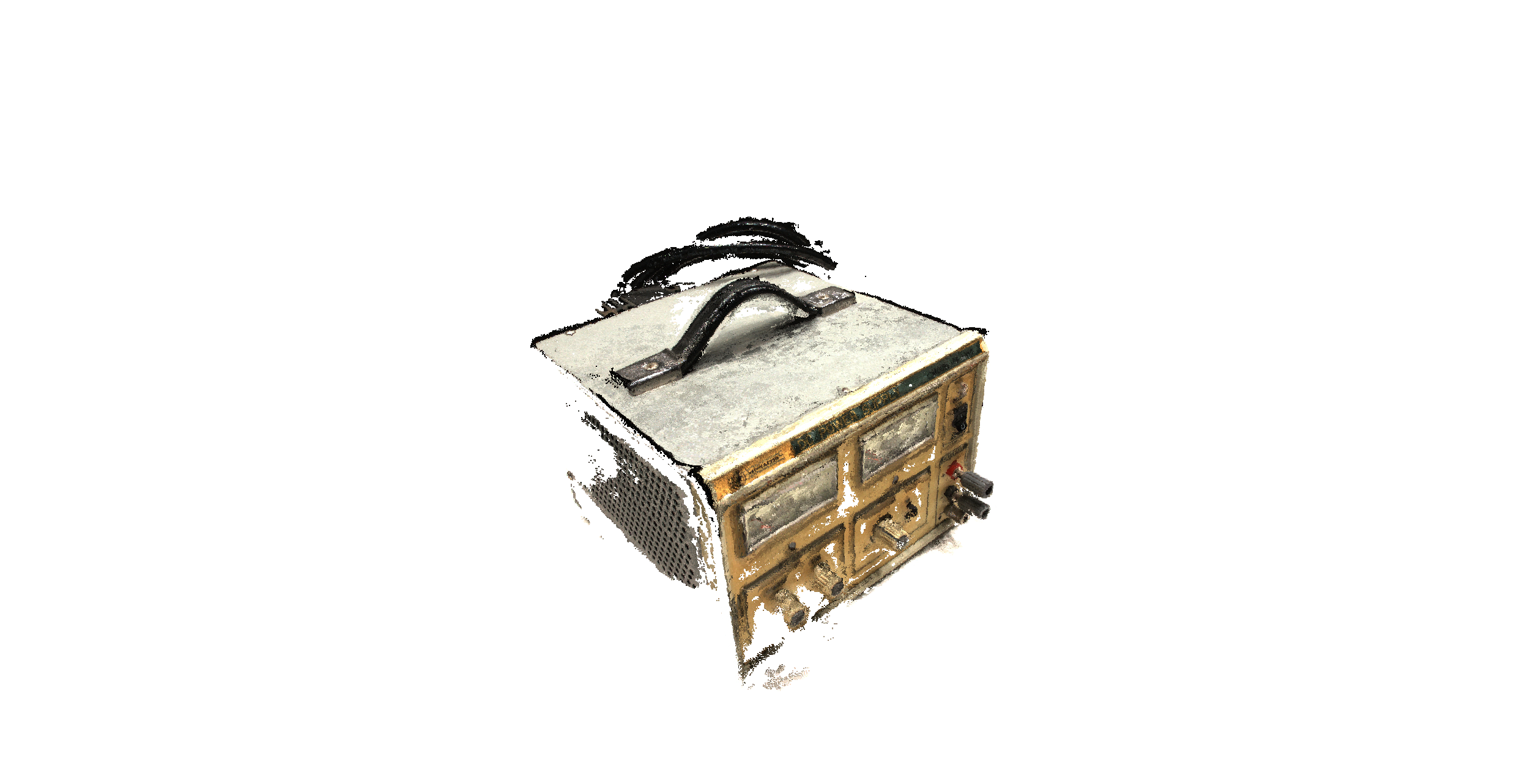} \\
			
		\end{tabular}
	}
	\caption{Point cloud reconstructions. From left to right: ground truth, MVSNet, SurfaceNet and ours. Our reconstruction results provide a better completeness than SurfaceNet and appear similar to the supervised MVSNet results.  }
	\label{tbl:table_of_figures}
\end{figure}

\begin{table}[tb]
	\centering
	\resizebox{\columnwidth}{!}{
		\begin{tabular} {ccccccc}
			\toprule
			method & acc. & comp. & over. F& prec. & rec. & over. F \\ 
			\cmidrule(lr){2-4} \cmidrule(lr){5-7}
			&  &   &  &   & (1 mm) in \% &  \\
			\midrule
			Self DTU(d=128)                & 0.881           & 1.073           & 0.977    & 61.54 & 44.98   & 51.98\\
			Self DTU (d=256)                    & 1.159       & \textbf{0.6083}       & 0.8837  & 64.85 & \textbf{64.68} & 63.57   \\
			Self PT bMVS, Self FT DTU                    & 0.9448           & 0.6345         & 0.7896     & 68.43 & 63.38 & 64.42   \\
			Sup PT bMVS, Self FT DTU                    & 0.7808         & 0.6769        & 0.7288    & 74.54 & 64.35 & 67.49   \\
			
			Ours (Meta PT bMVS, Self FT DTU, no conf. mask)                    & 0.7242           & 0.8422           & 0.7832    & 75.22 & 60.25 & 65.31 \\
			Ours (Meta PT bMVS, Self FT DTU)                                  & \textbf{0.5942} & 0.7787 & \textbf{0.6865}  &\textbf{80.18} & 63.58 & \textbf{68.67}\\ \bottomrule
		\end{tabular}
		
	}
	\caption{Ablation study (bold shows best results). Acronyms follow Table~\ref{tab:results1}. Our meta learning approach achieves better overall scores than the other training variants. }
	\label{tab:table_of_ablation}
\end{table}

\subsection{Qualitative Results}
Figure \ref{tbl:table_of_figures} display the qualitative evaluation of our proposed method with respect to supervised methods (~\citep{MVSNet,Surfacenet}). Our method provides a superior completeness and as it can be observed from the reconstruction, some surrounding structures are also reconstructed which are not present in the ground truth.

\subsection{Ablation Studies}
We perform ablation studies on the following training conditions:
\begin{itemize}[noitemsep,topsep=0pt]
  \item Self-supervised MVSNet setup (\textit{Self DTU (d=256)}) similar to~\citep{khot}, with twice the depth discretization level (d=256). It was trained on DTU train split, and has different loss hyperparameters (such as reprojection loss weights as proposed in ~\citep{khot}). 
  \item Similar as the previous setup, but pre-trained (PT) on BlendedMVS using self-supervi\-sed learning (\textit{(Self PT bMVS, Self FT DTU)}) and supervised learning (\textit{Sup PT bMVS, Self FT DTU)}). The model is fine-tuned (FT) on DTU using self-supervised learning.
  \item Our meta-training setup without the confidence mask training (\textit{Ours (Meta PT bMVS, Self FT DTU, no conf. mask)}).
  \item Our proposed meta-training setup with the confidence mask training (\textit{Ours (Meta PT bMVS, Self FT DTU)}).
\end{itemize}

Table \ref{tab:table_of_ablation} shows the results for these variations of our model. The \textit{overall} scores highlight that meta learning outperforms the straightforward fine-tuning strategy (PT bMVS) with the same sequence of datasets, even if it is pre-trained supervised. Confidence weight masks are effective for decreasing the effect of outliers during learning which improves performance.

\section{Conclusions}
Adaptability to new domains through self-supervision is a powerful property, especially for a multi-view stereo learning module where dense ground-truth depth data is tedious and difficult to obtain.
We propose a meta learning approach which trains a network for self-supervised adaptation to a novel data domain with changes in environment and conditions. 
Our approach learns a loss confidence mask for self-supervised learning. 
In our experiments, we demonstrate that our meta-learning helps to train the network for adapting to new domains using self-supervision.
Our approach can improve self-supervised domain adaptation performance over naive pre-training using depth supervision.
It achieves reconstruction results which well compare with a previous supervised method and classical methods, and can improve performance over a self-supervised baseline. 

Meta learning and multi-view stereo learning is a popular topic in the field of machine learning and computer vision. In the future, we will investigate architectures for high resolution images for an improved and more detailed reconstruction. 

\section*{Acknowledgements}
This work has been partially funded by the Deutsche Forschungsgemeinschaft (DFG, German Research Foundation) under Germany's Excellence Strategy - EXC number 2064/1 - project number 390727645 and SFB 1233 - project number 276693517. It was supported by the German Federal Ministry of Education and Research (BMBF): T\"ubingen AI Center, FKZ: 01IS18039A and Cyber Valley. The authors thank the International Max Planck Research School for Intelligent Systems (IMPRS-IS) for supporting Arijit Mallick.

\bibliography{egbib}

\clearpage
\begin{huge}
\textbf{\LARGE Supplemental Material}
\end{huge}

\setcounter{page}{1}
\setcounter{section}{0}
\makeatletter

\section{Introduction}
In the following we provide additional details and results for our approach.

\section{Network Architecture}

\begin{figure}
  \centering
    \includegraphics[width =1\textwidth]{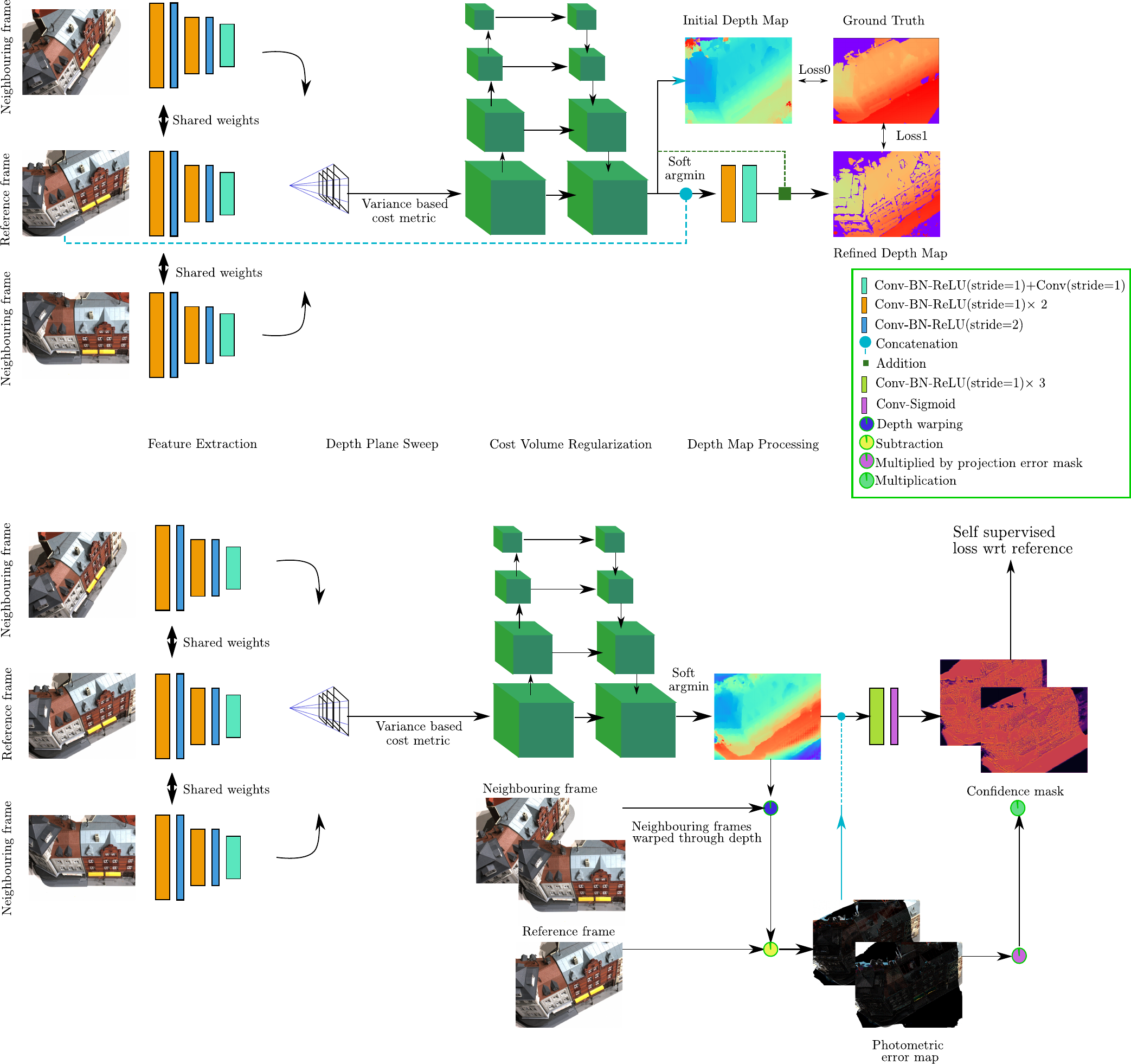}
    \caption{Network architecture difference between MVSNet (top) and our model (bottom). Our model builds on the initial stages of MVSNet: Deep features are extracted from the reference frame and the neighbouring frames. A plane-sweep cost volume is determined by homographic warping of the neighbouring feature maps to the reference view in a set of depth planes. This cost volume is refined in an encoder-decoder architecture and a depth map is obtained using a soft argmin operation. In case of MVSNet, this initial depth map is further improved by a refinement network. Supervised losses are determined that compare the refined depth with ground truth. In our model, we do not use the refinement branch. Instead, we determine photometric error maps by warping the neighbouring frames to the reference view and comparing them with the reference image. These photometric error maps are input to a confidence mask prediction network which also receives out-of-image projection masks. Finally, a self-supervised loss is computed by utilising the confidence mask on the photometric error maps. }
    \label{fig:struct}
\end{figure}

We base our network architecture on MVSNet~\citep{MVSNet}.
We do not use the depth refinement module, but extend the network with a subnetwork which predicts a confidence mask for the self-supervised loss. 
We provide a comparison of the two architectures in Fig.~\ref{fig:struct}.
The confidence mask subnetwork is a 4-layer CNN with a sigmoid activation unit at the end to generate values between 0 to 1. The confidence mask prediction network comprises of a combination of two basic sub-blocks. The first sub-block consists of a 2D convolutional layer (kernel size=3, stride=1) followed by a BatchNorm layer and ReLU as it's activation function. This sub-block layer is used 3 times succesively and then is followed by a final sub-block which consists of a 2D convolutional layer (kernel size=3, stride=1) followed by sigmoid activation function.
The subnetwork receives as input the out-of-image projection masks and a photometric error maps for each neighbouring view.
The photometric error maps are determined by warping the neighbouring views to the refence view and taking the difference.

\begin{figure}
\centering
\resizebox{\columnwidth}{!}{
\begin{tabular}{cccccc}
   
     reference frame & confidence mask 1 & neighbouring frame 1 & confidence mask 2 & neighbouring frame 2 & depth \\
    
    \includegraphics[width=.16\textwidth]{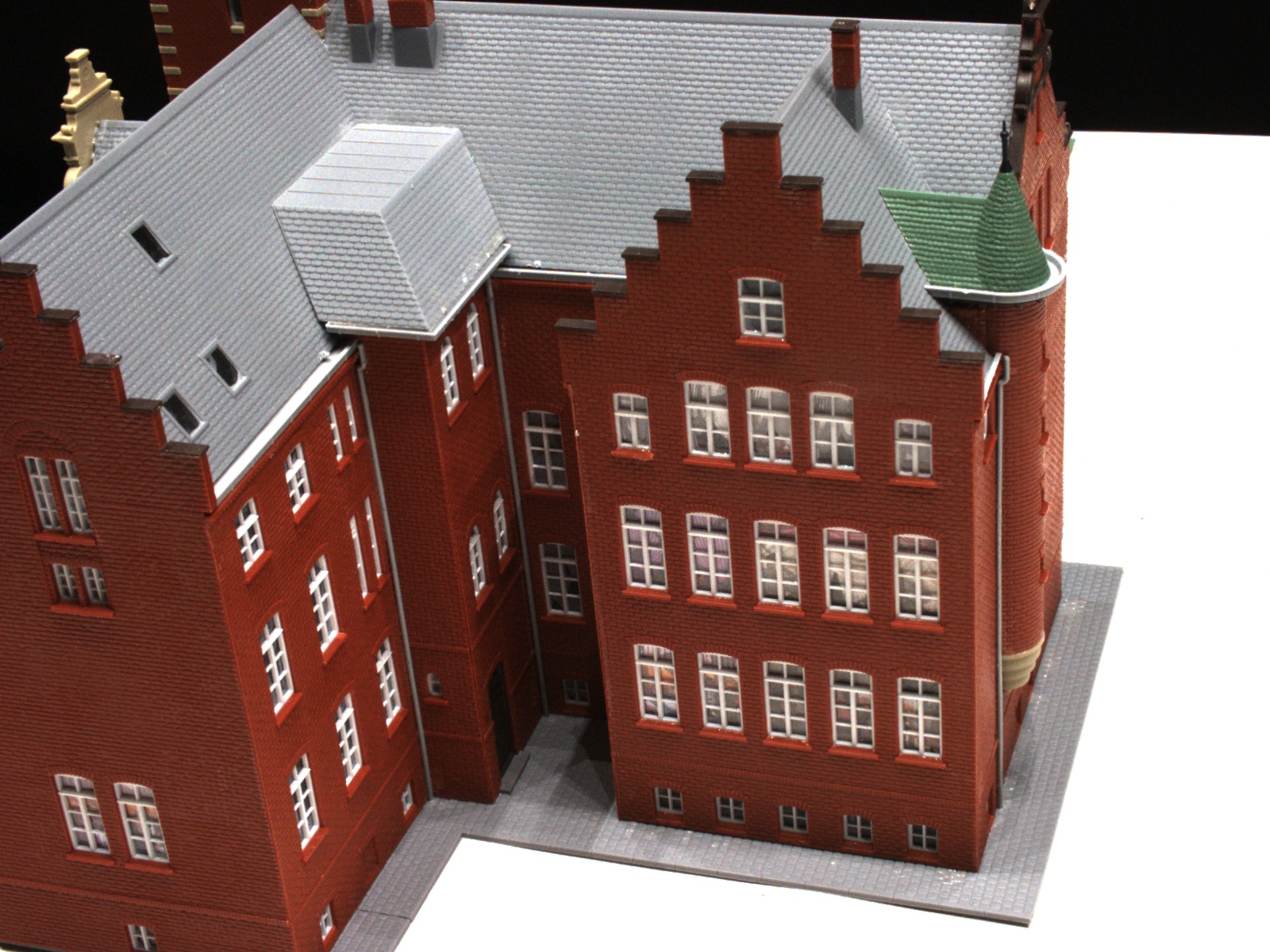} & 
    \includegraphics[width=.16\textwidth]{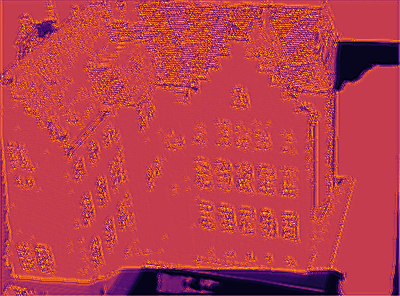} & 
    \includegraphics[width=.16\textwidth]{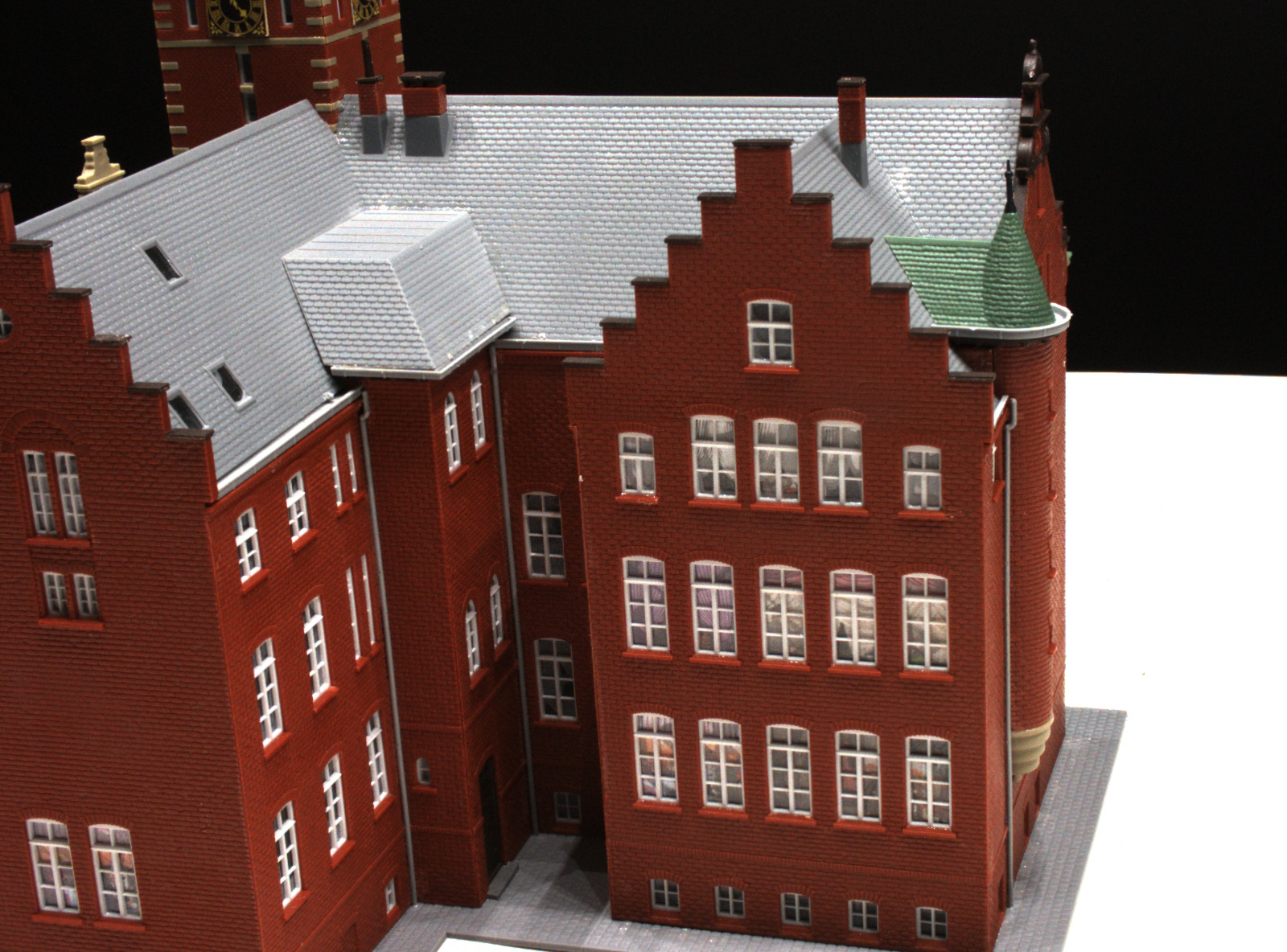} & 
    \includegraphics[width=.16\textwidth]{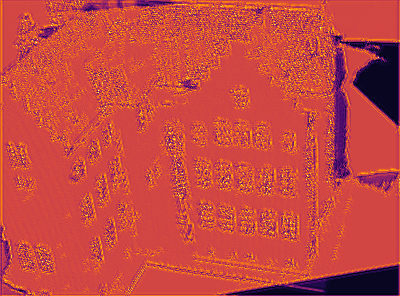} & 
    \includegraphics[width=.16\textwidth]{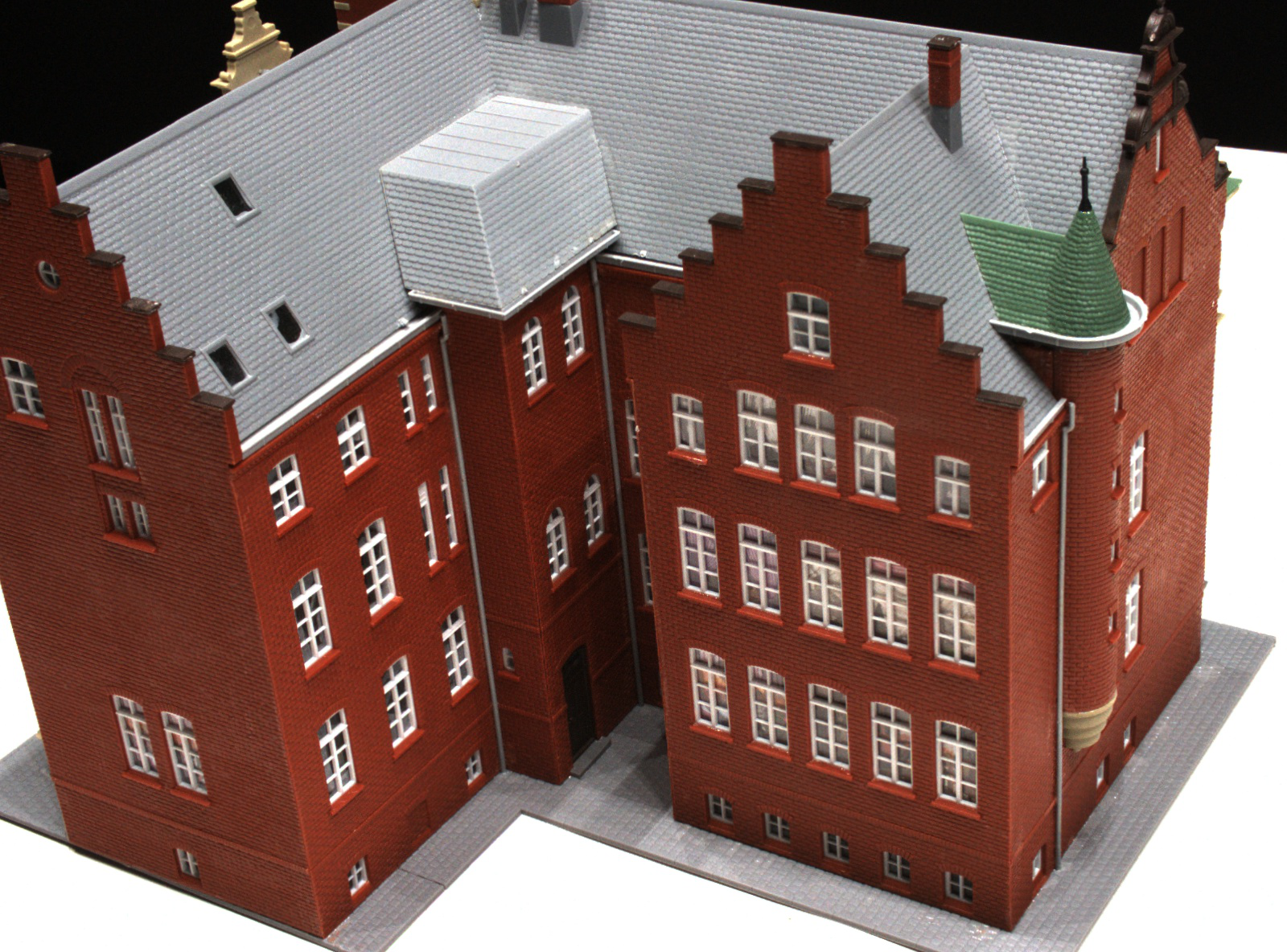} & 
    \includegraphics[width=.16\textwidth]{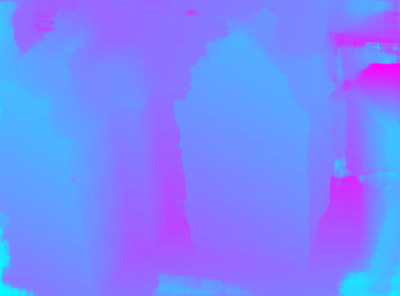} \\
    
    \includegraphics[width=.16\textwidth]{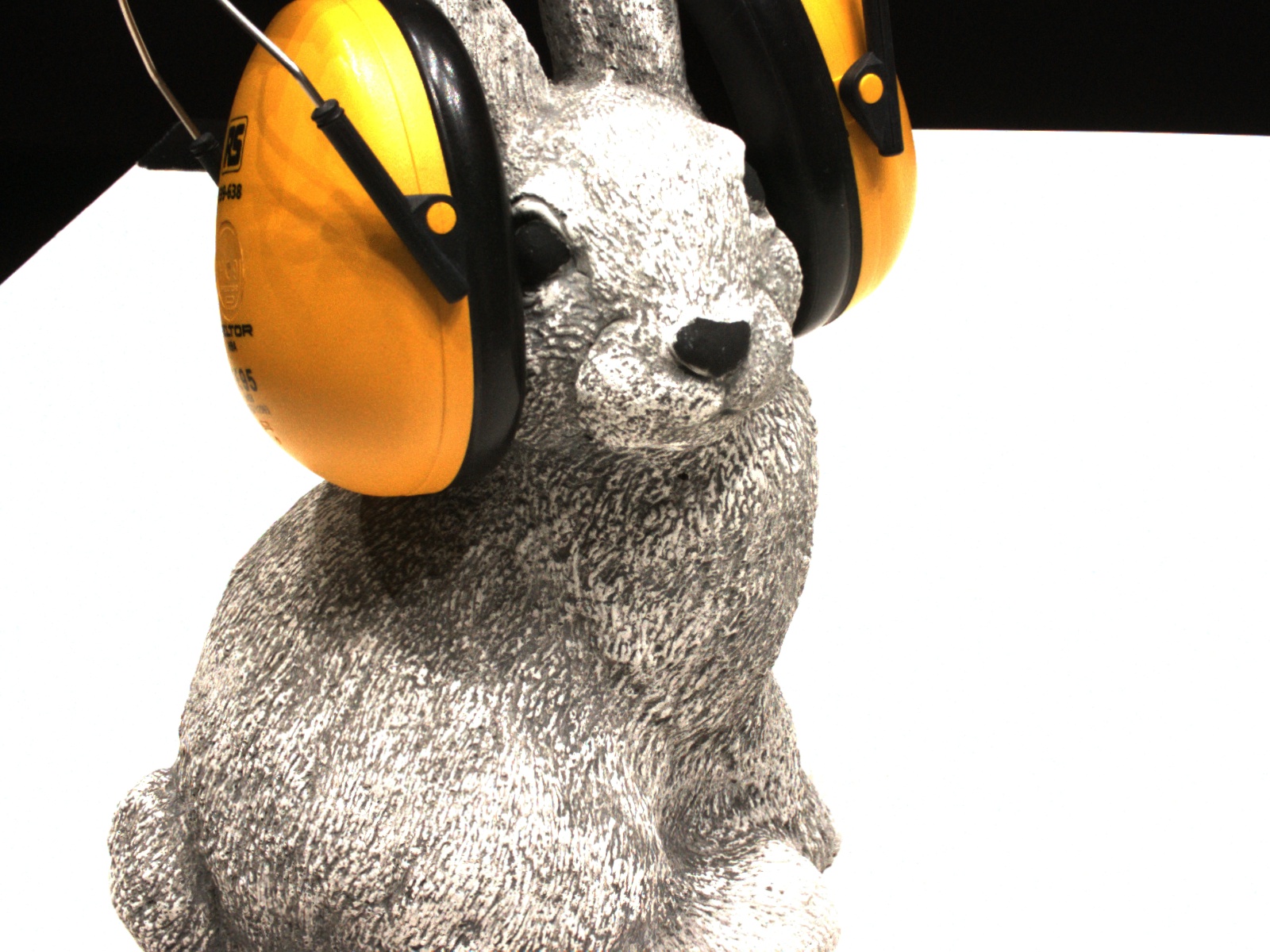} & 
    \includegraphics[width=.16\textwidth]{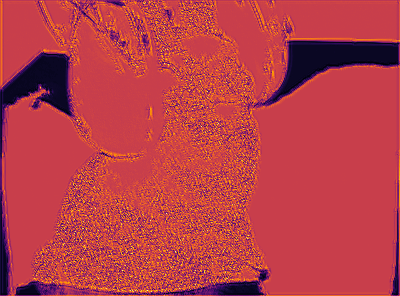} & 
    \includegraphics[width=.16\textwidth]{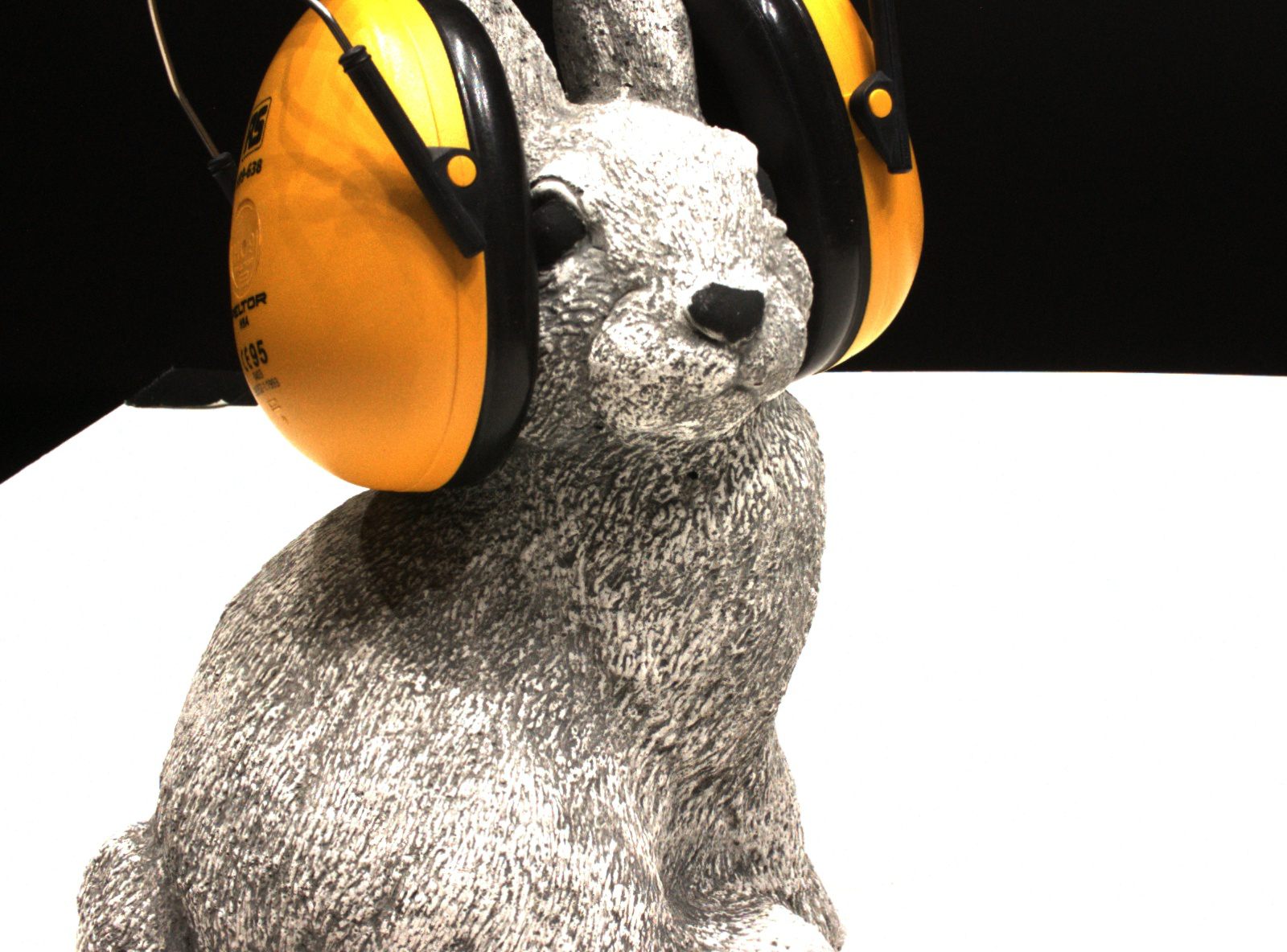} & 
    \includegraphics[width=.16\textwidth]{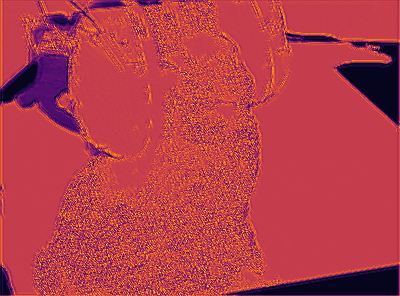} & 
    \includegraphics[width=.16\textwidth]{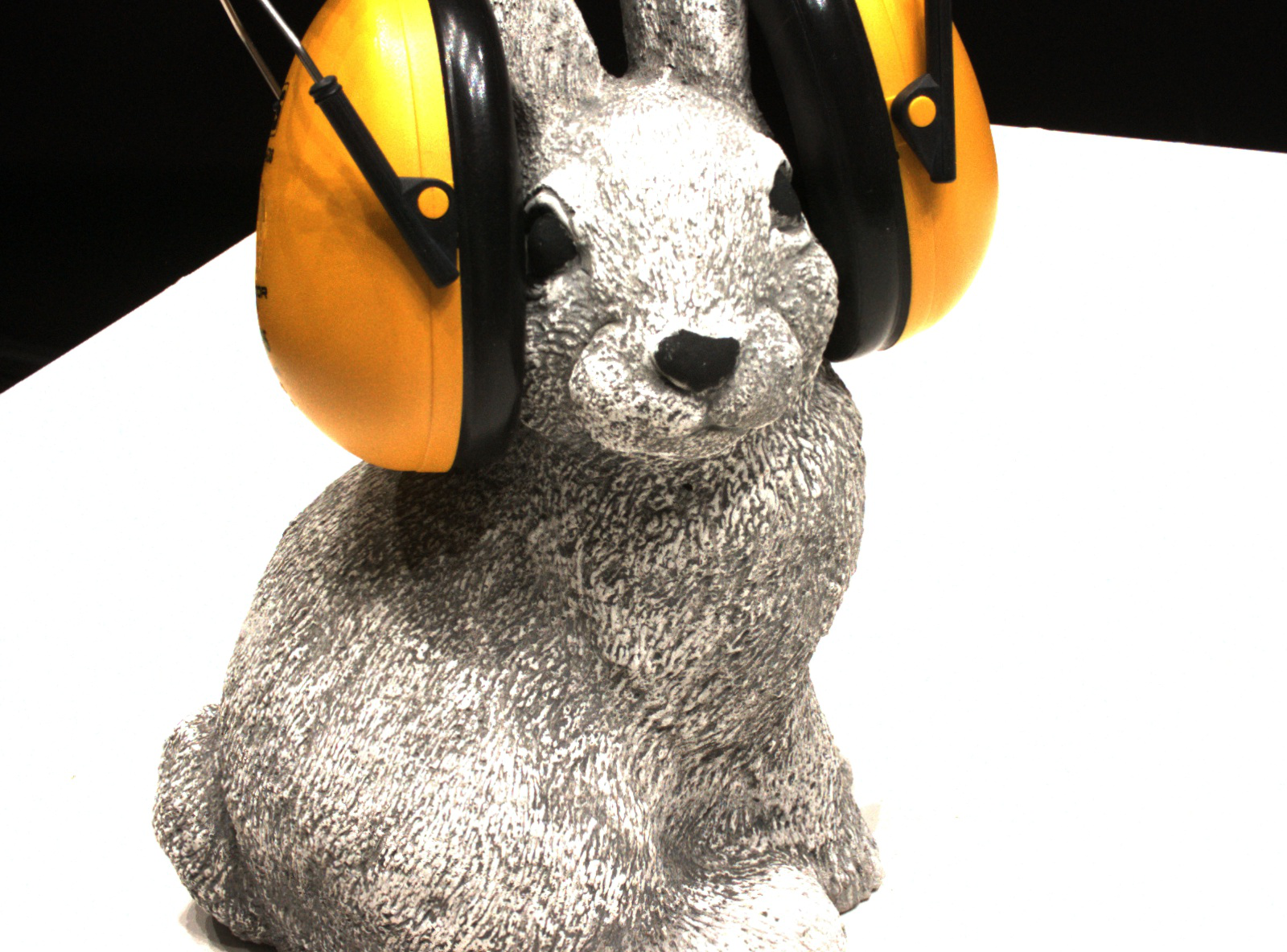} & 
    \includegraphics[width=.16\textwidth]{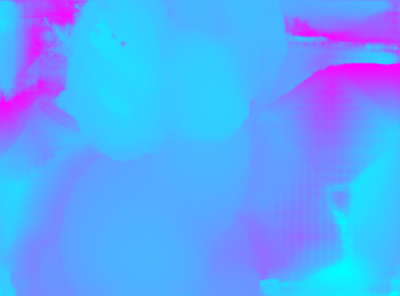} \\

\end{tabular}
}
\caption{Examples of predicted confidence masks. From left to right: reference frame, predicted confidence mask for first view (red: 1, black: 0 confidence), first view, predicted confidence mask for second view, second view, predicted depth maps on the DTU dataset. }
\label{tbl:conf}
\end{figure}

\section{Additional Quantitative Results}
We also provide evaluation results on the DTU Buddha scan (see Table~\ref{tab:eval}).
Results of several classical and supervised methods are taken from~\citep{Paschalidou_2018_CVPR}. 
Supervised MVSNet~\citep{MVSNet} fares best, while our self-supervised method ranks second and outperforms supervised and classical methods, highlighting the efficacy of our meta-learning approach.

\begin{table}[htbp]
\centering
\begin{tabular}{ lc c c }
\toprule
method              & accuracy & completeness & overall  \\
\midrule

MVSNet~\citep{MVSNet} (Sup DTU)         & \bf 0.234 & \bf 0.278 & \bf 0.257   \\
Ours best (Meta PT bMVS, Sup FT DTU )      & 0.455 & 0.335 &  0.395\\
Ours (no mask)  (Meta PT bMVS, Sup FT DTU )    & 0.483 & 0.339 & 0.412  \\
SurfaceNet~\citep{Surfacenet} (Sup DTU, from~\citep{Paschalidou_2018_CVPR}) & 0.738 & 0.677 & 0.707  \\
Hartmann et al.~\citep{Hartmann}(Sup DTU, from~\citep{Paschalidou_2018_CVPR})    & 0.637 & 1.057 & 0.847    \\
RayNet~\citep{Paschalidou_2018_CVPR} (Sup DTU, from~\citep{Paschalidou_2018_CVPR})          & 1.993 & 0.481 & 1.237   \\
Ulusoy et al.~\citep{Ul} (C, from~\citep{Paschalidou_2018_CVPR})         & 4.784 & 0.953 & 2.868   \\
ZNCC~\citep{classic}(C, from~\citep{Paschalidou_2018_CVPR})           & 6.107 & 0.646 & 3.376   \\
SAD~\citep{classic}(C, from~\citep{Paschalidou_2018_CVPR})            & 6.683 & 0.753 & 3.718   \\
\bottomrule

\end{tabular}
\caption{Ranking of several methods by overall metric on the DTU Buddha dataset. Lower is better (best as bold). C: classical, Sup: supervised, Self: self-supervised, Meta: meta-learning. Our self-supervised meta-learning approach performs better than several supervised and classical methods. PT bMVS, FT DTU denotes pre-trained with Blended MVS dataset and fine-tuned with DTU dataset. DTU denotes trained on DTU dataset.}
\label{tab:eval}
\end{table}

\begin{figure}
\centering
\resizebox{\columnwidth}{!}{
\begin{tabular}{cccc}
   
     Ground truth & MVSNet~\citep{MVSNet} & SurfaceNet\citep{Surfacenet} & Ours (Meta, Self) \\
    
    \includegraphics[trim={20cm 8cm 17cm 3cm},clip,width=.23\textwidth,height = 0.17\textwidth]{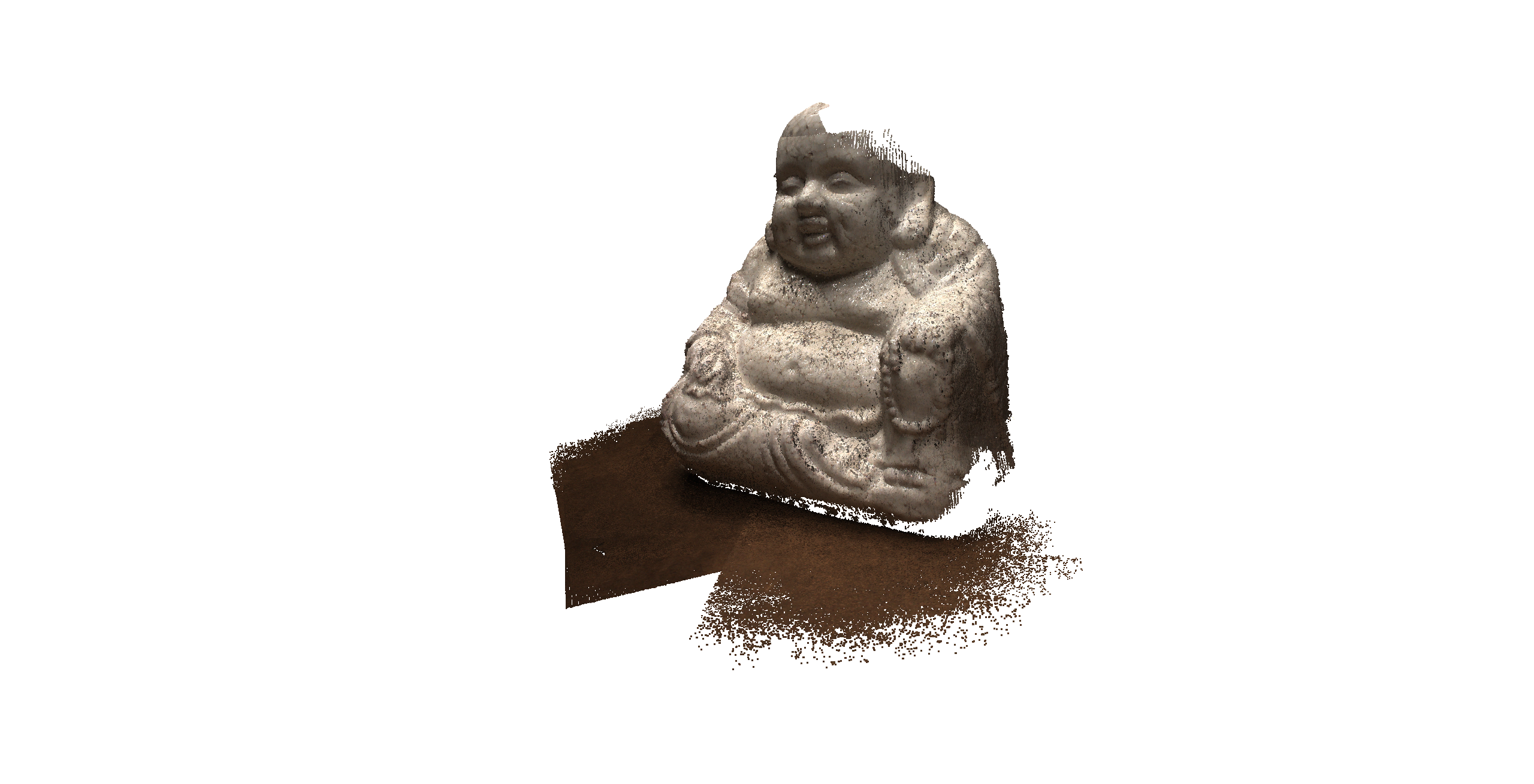} & 
    \includegraphics[trim={20cm 8cm 17cm 3cm},clip,width=.23\textwidth,height = 0.17\textwidth]{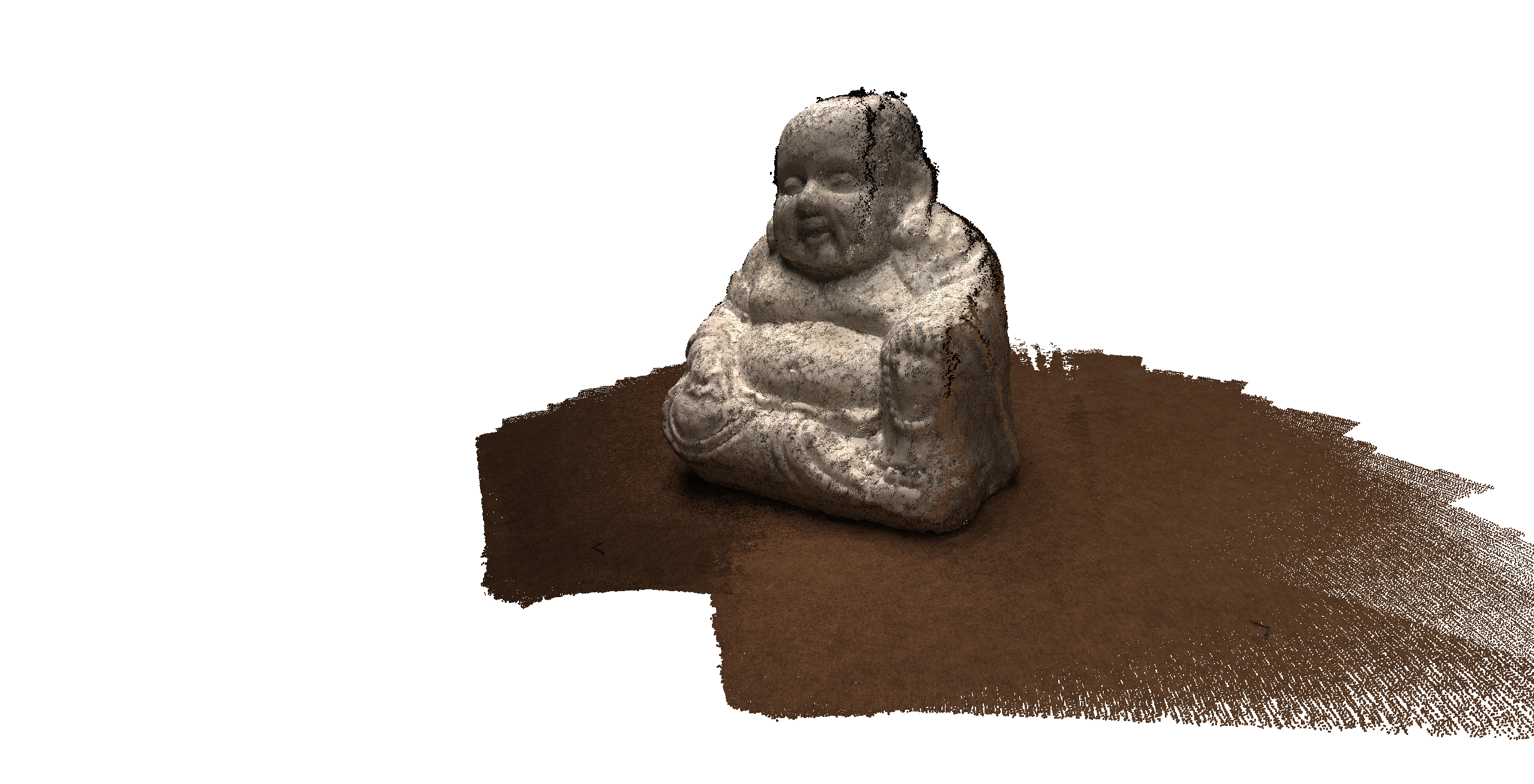} & 
    \includegraphics[trim={20cm 8cm 17cm 3cm},clip,width=.23\textwidth,height = 0.17\textwidth]{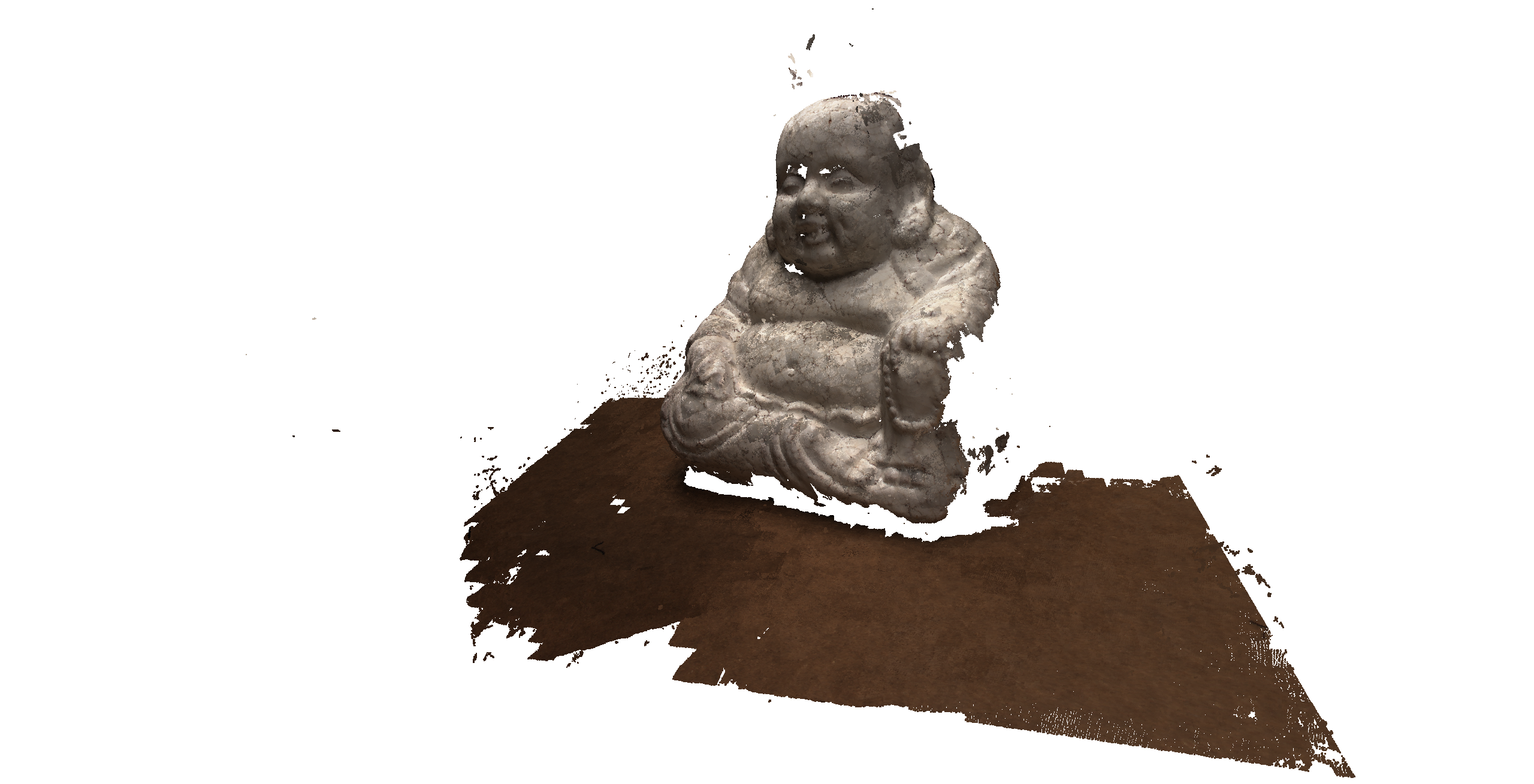} & 
    \includegraphics[trim={20cm 8cm 17cm 3cm},clip,width=.23\textwidth,height = 0.17\textwidth]{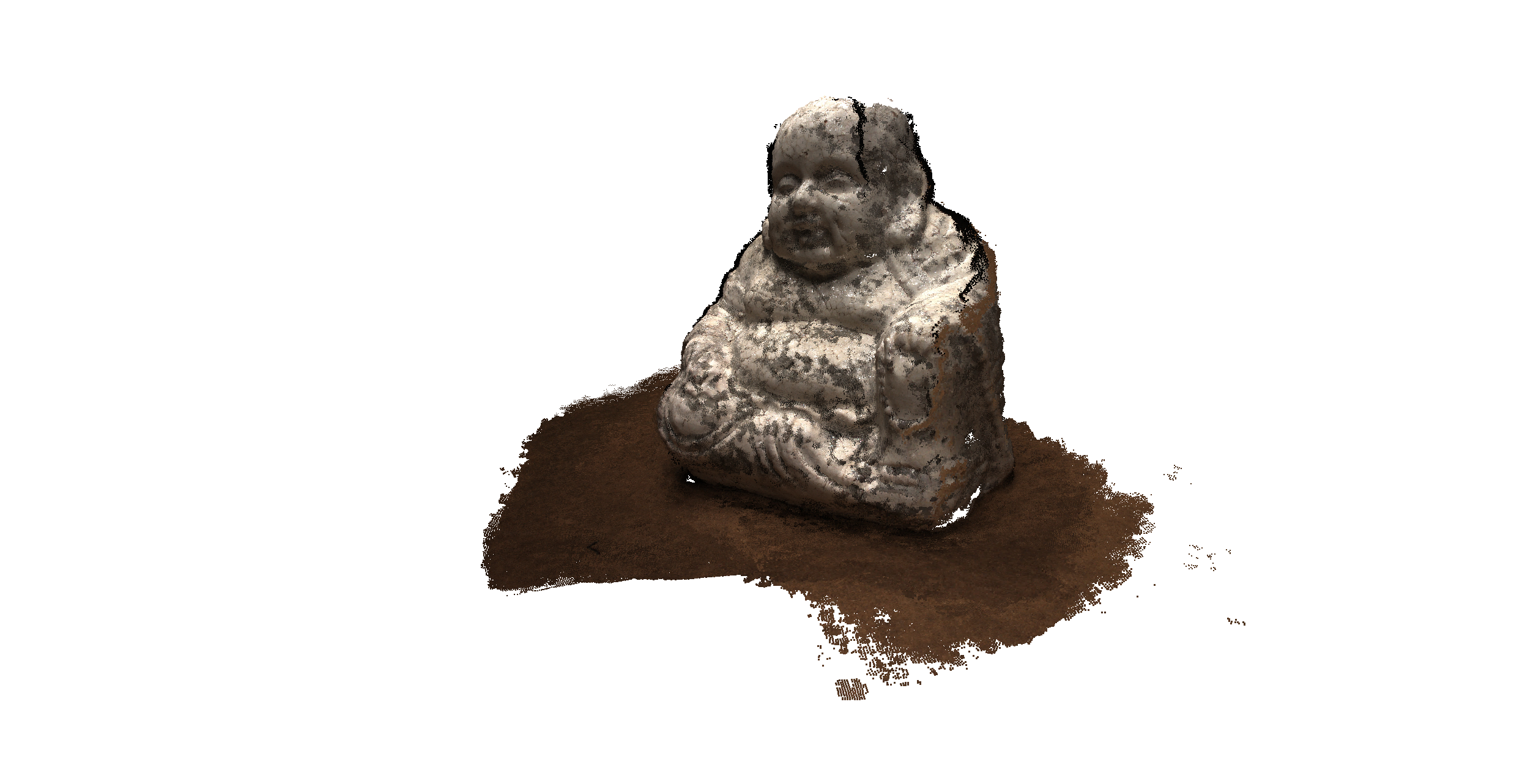} \\

\end{tabular}
}
\caption{Point-cloud reconstruction results on the DTU Buddha dataset. The qualitative results of our meta-learning approach appear superior to supervised SurfaceNet~\citep{Surfacenet} and fairly close to MVSNet~\cite{MVSNet}. }
\label{tbl:refe}
\end{figure}

\section{Additional Qualitative Results}

\begin{figure}
\centering     
\subfigure[Lake-side]{\label{fig:a}\includegraphics[width=60mm,height = 0.35\textwidth]{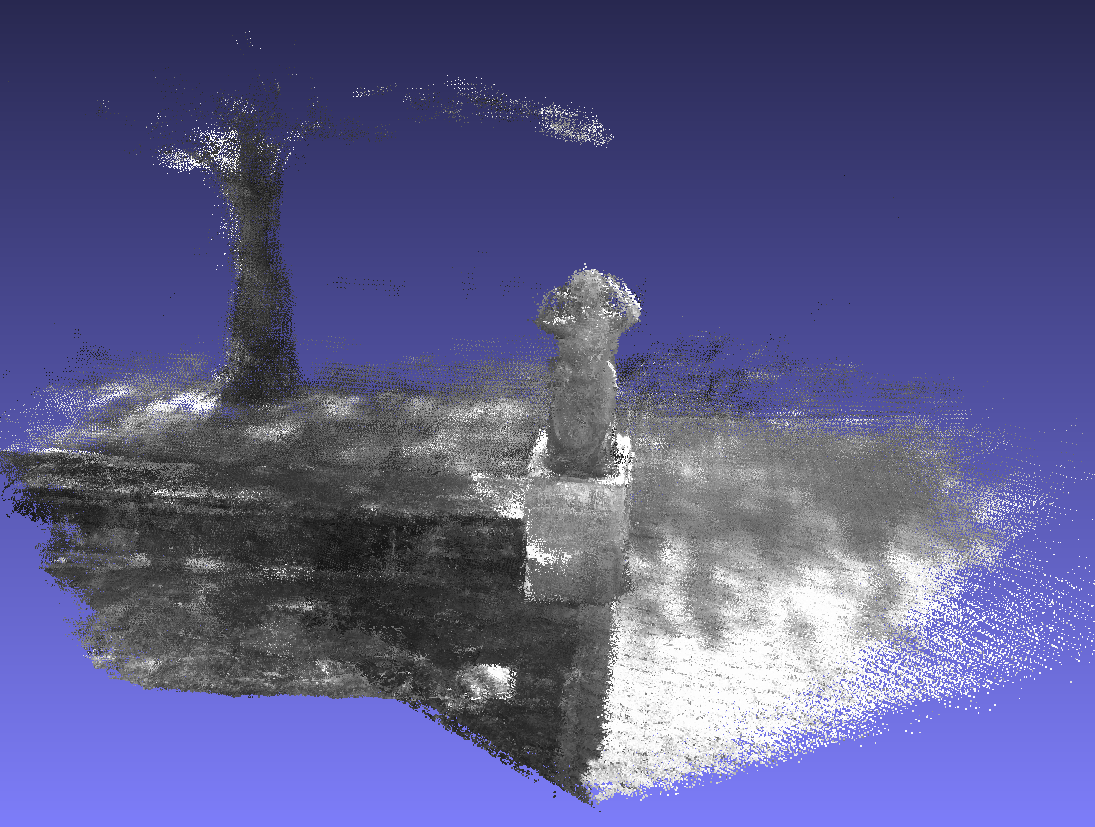}}
\subfigure[Sandbox]{\label{fig:b}\includegraphics[width=60mm,height = 0.35\textwidth]{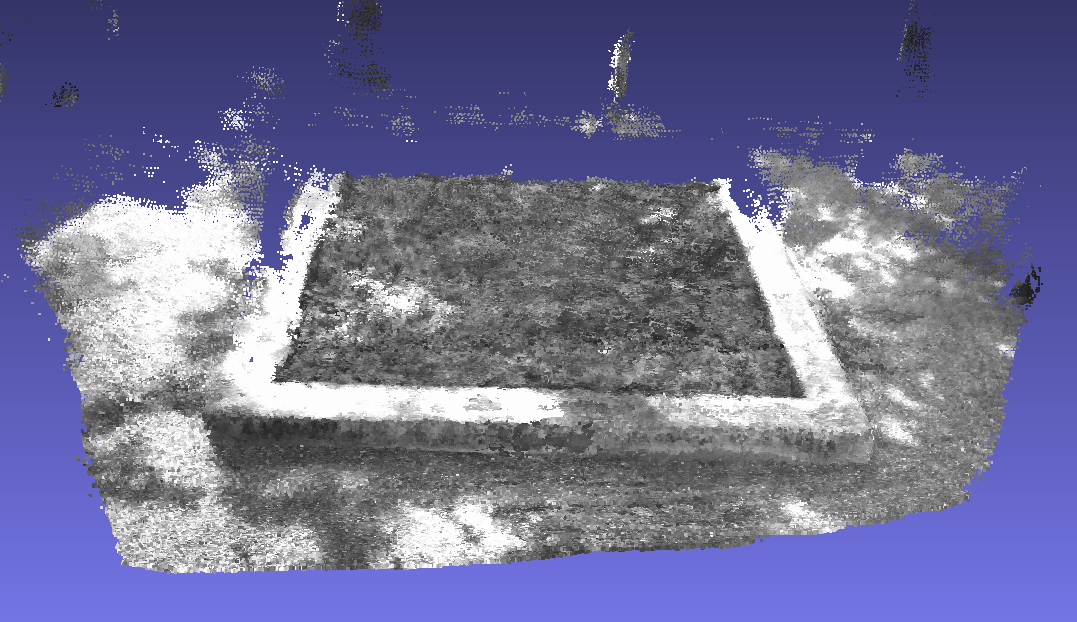}}
\caption{Point-cloud reconstruction results of our meta-learning approach on ETH3D low resolution multiview dataset reconstruction.}
\label{fig:ETH3D}
\end{figure}

\paragraph{ETH3D.} 
Qualitative evaluation was performed on the ETH3D test dataset for low resolution multiview stereo. The model was meta-trained on BlendedMVS and fine-tuned on ETH3D low resolution training dataset. The test point-clouds show clear reconstruction results (see Fig.~\ref{fig:ETH3D}).

\paragraph{DTU.} 
We provide additional point-cloud reconstruction results on the DTU dataset in Fig.~\ref{tbl:depth}.
We also show depth maps predicted by our approach in Fig.~\ref{tbl:conf}).

\begin{figure}
\centering
\resizebox{\columnwidth}{!}{
\begin{tabular}{|c|c|}
    \hline
    
    \includegraphics[trim={26cm 12cm 25cm 10cm},clip,width=.3\textwidth,height = 0.18\textwidth]{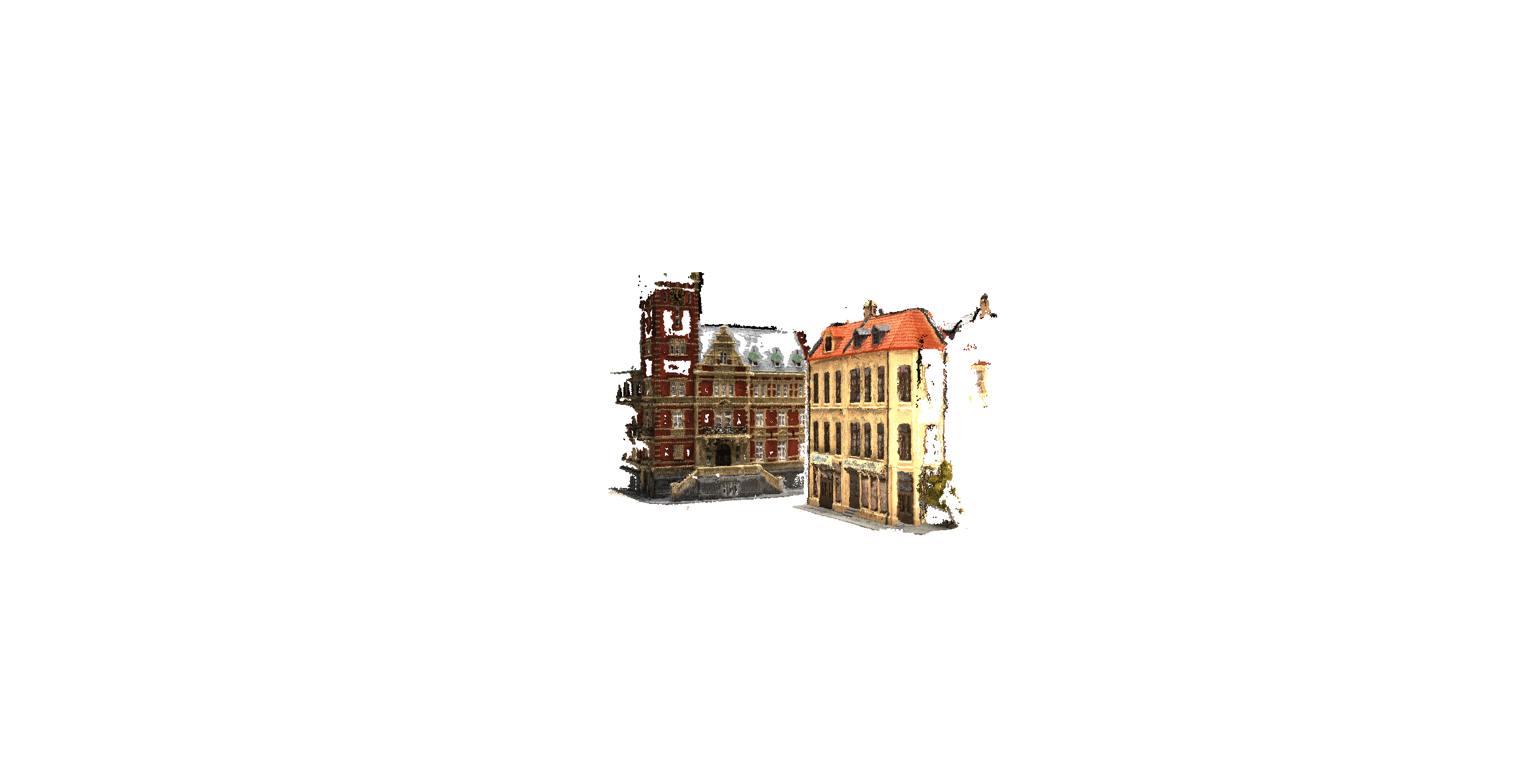} & 
    \includegraphics[trim={20cm 12cm 25cm 10cm},clip,width=.3\textwidth,height = 0.18\textwidth]{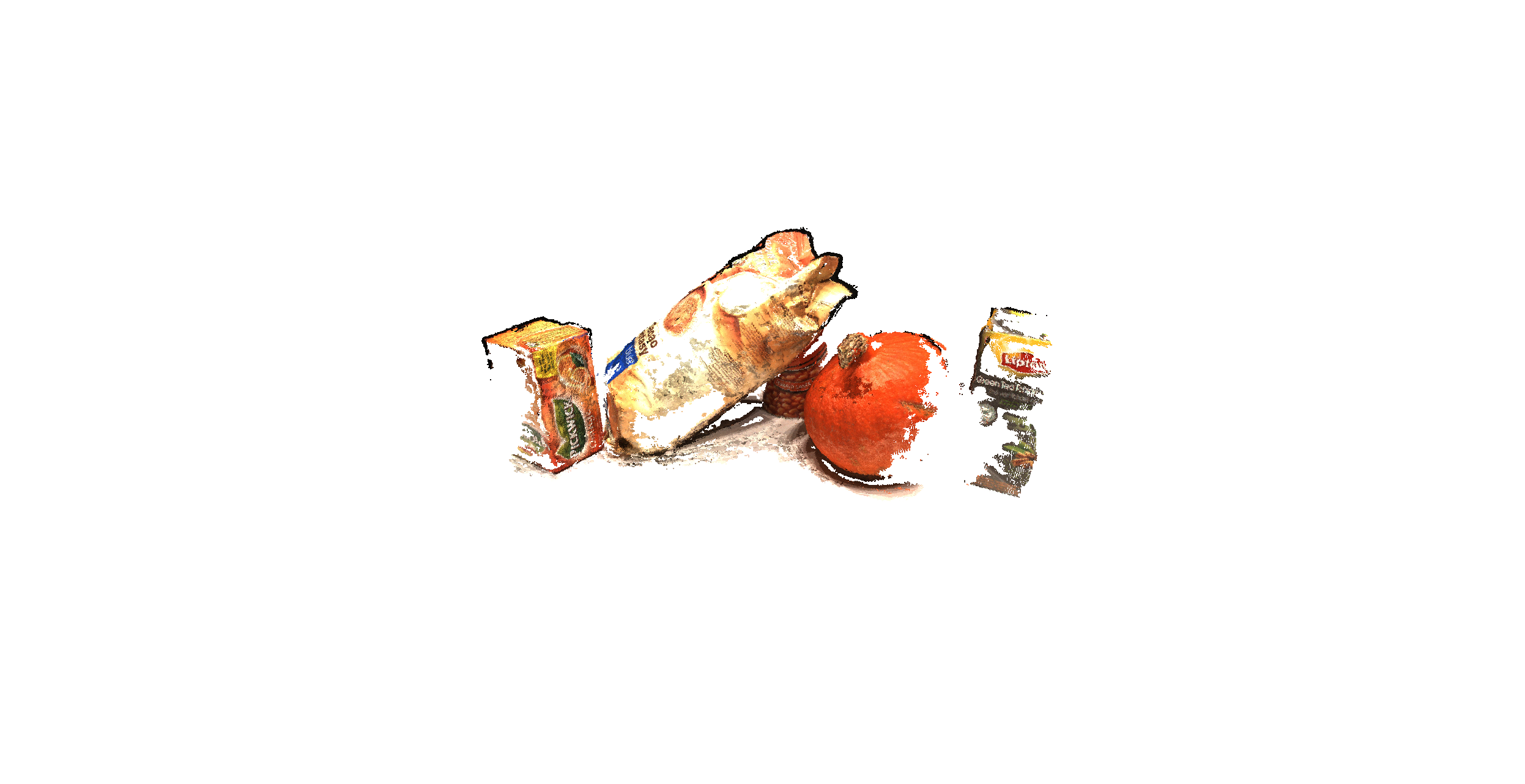} 
     \\
     \hline
     \includegraphics[trim={24cm 9cm 23cm 8cm},clip,width=.3\textwidth,height = 0.18\textwidth]{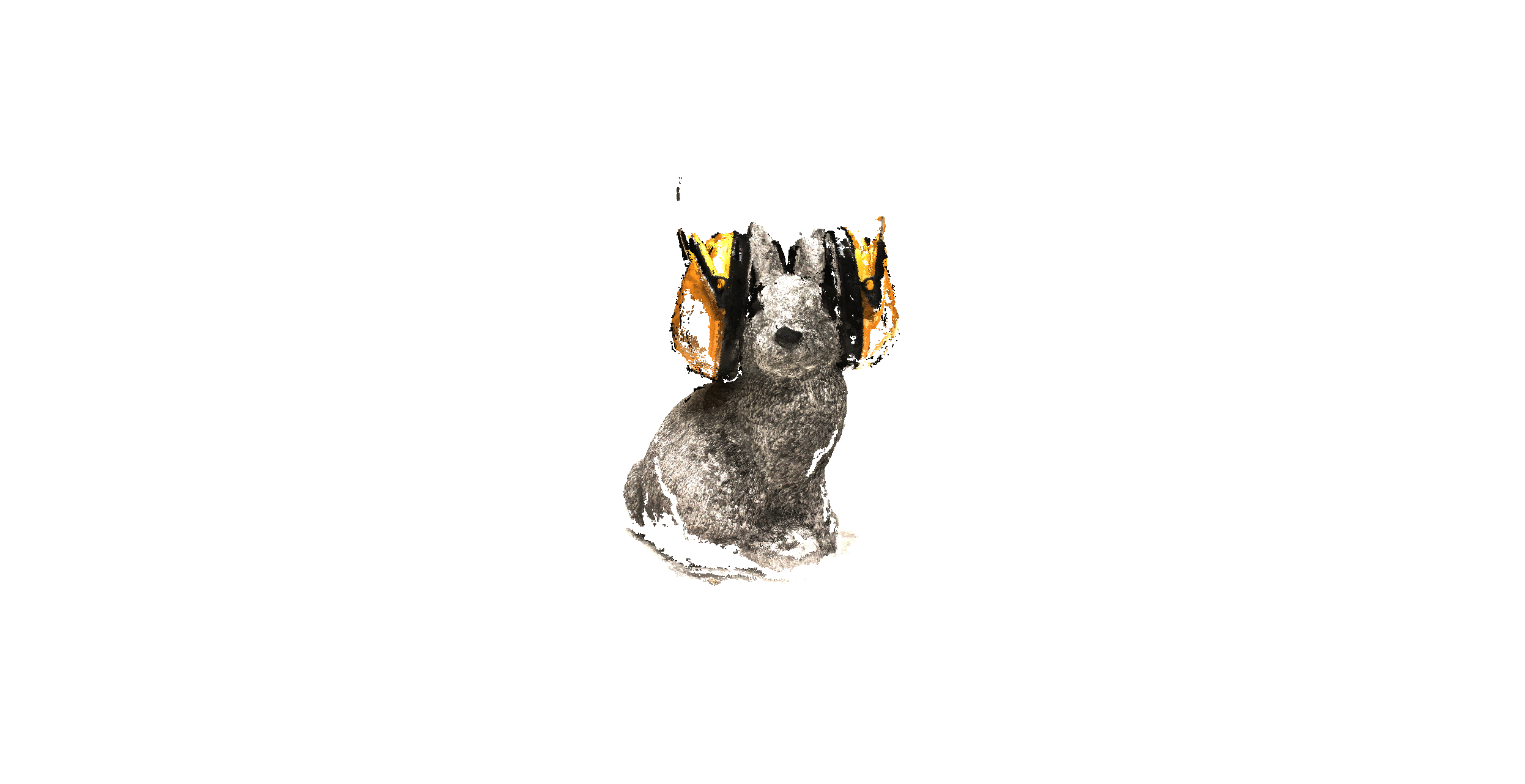} &  
    \includegraphics[trim={23cm 12cm 22cm 7cm},clip,width=.3\textwidth,height = 0.18\textwidth]{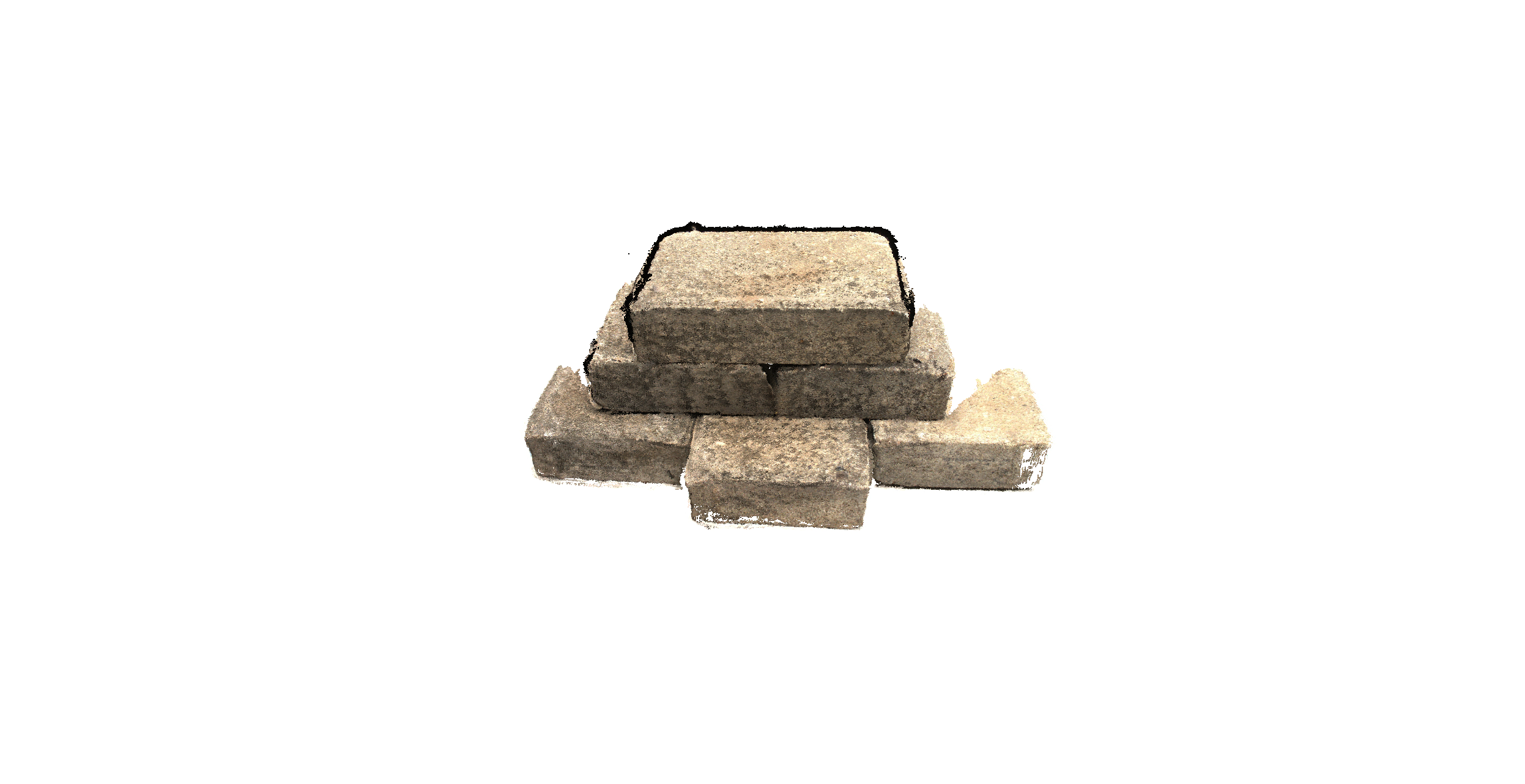}  \\
     \hline
     \includegraphics[trim={20cm 12cm 25cm 10cm},clip,width=.3\textwidth,height = 0.18\textwidth]{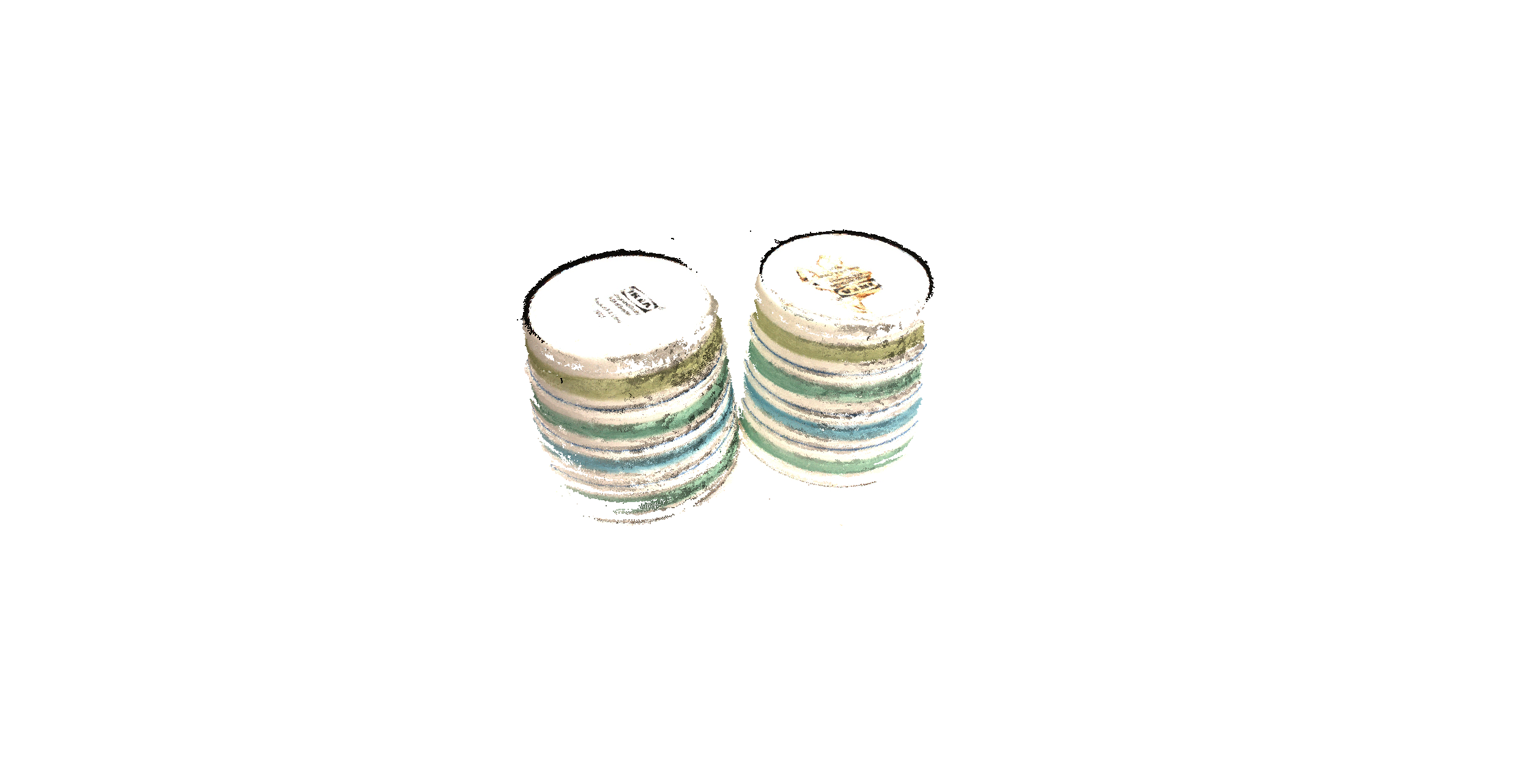} & 
    \includegraphics[trim={22cm 8cm 20cm 10cm},clip,width=.3\textwidth,height = 0.18\textwidth]{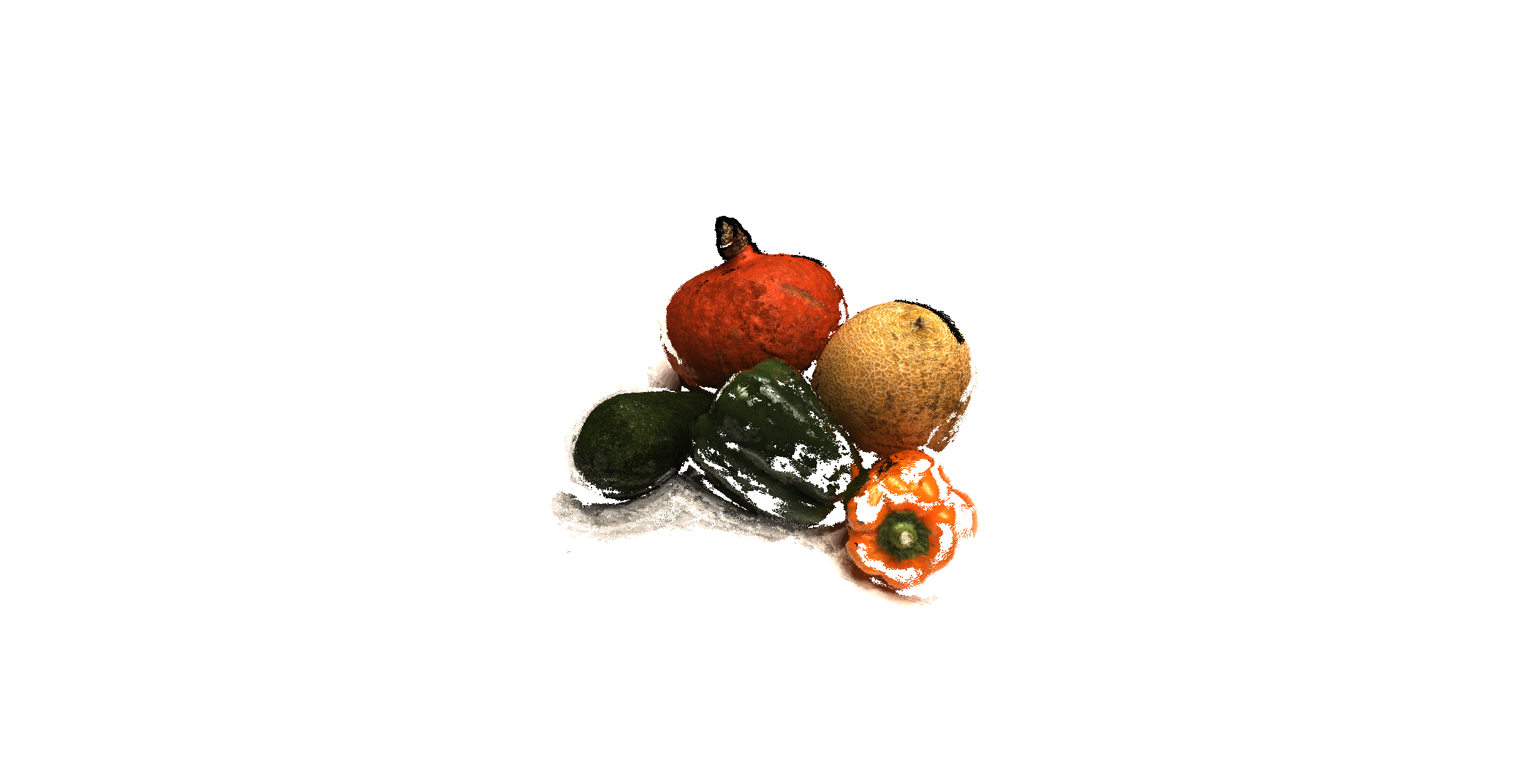}  \\
     \hline
    \includegraphics[trim={25cm 12cm 20cm 10cm},clip,width=.3\textwidth,height = 0.18\textwidth]{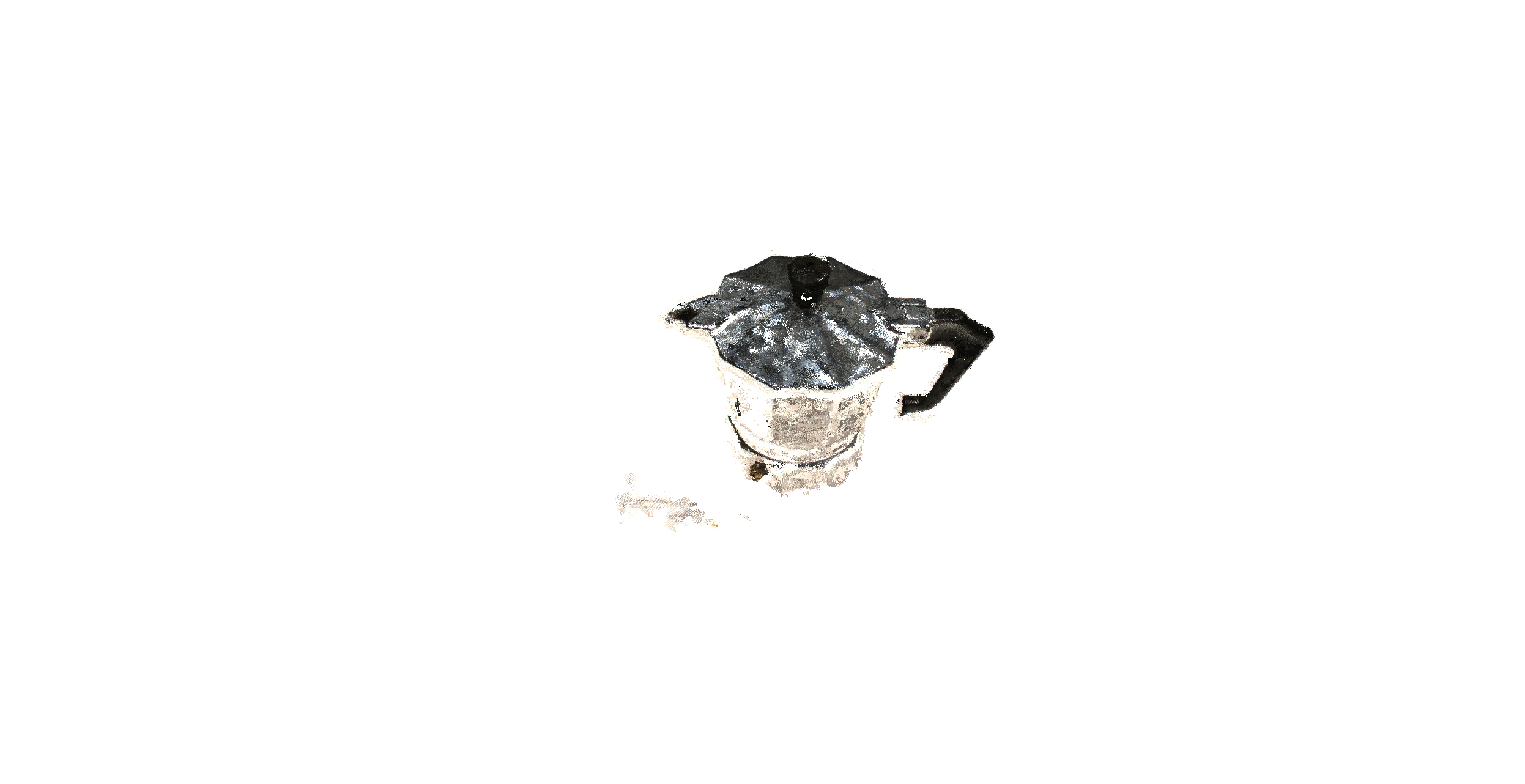} & 
    \includegraphics[trim={23cm 12cm 22cm 7cm},clip,width=.3\textwidth,height = 0.18\textwidth]{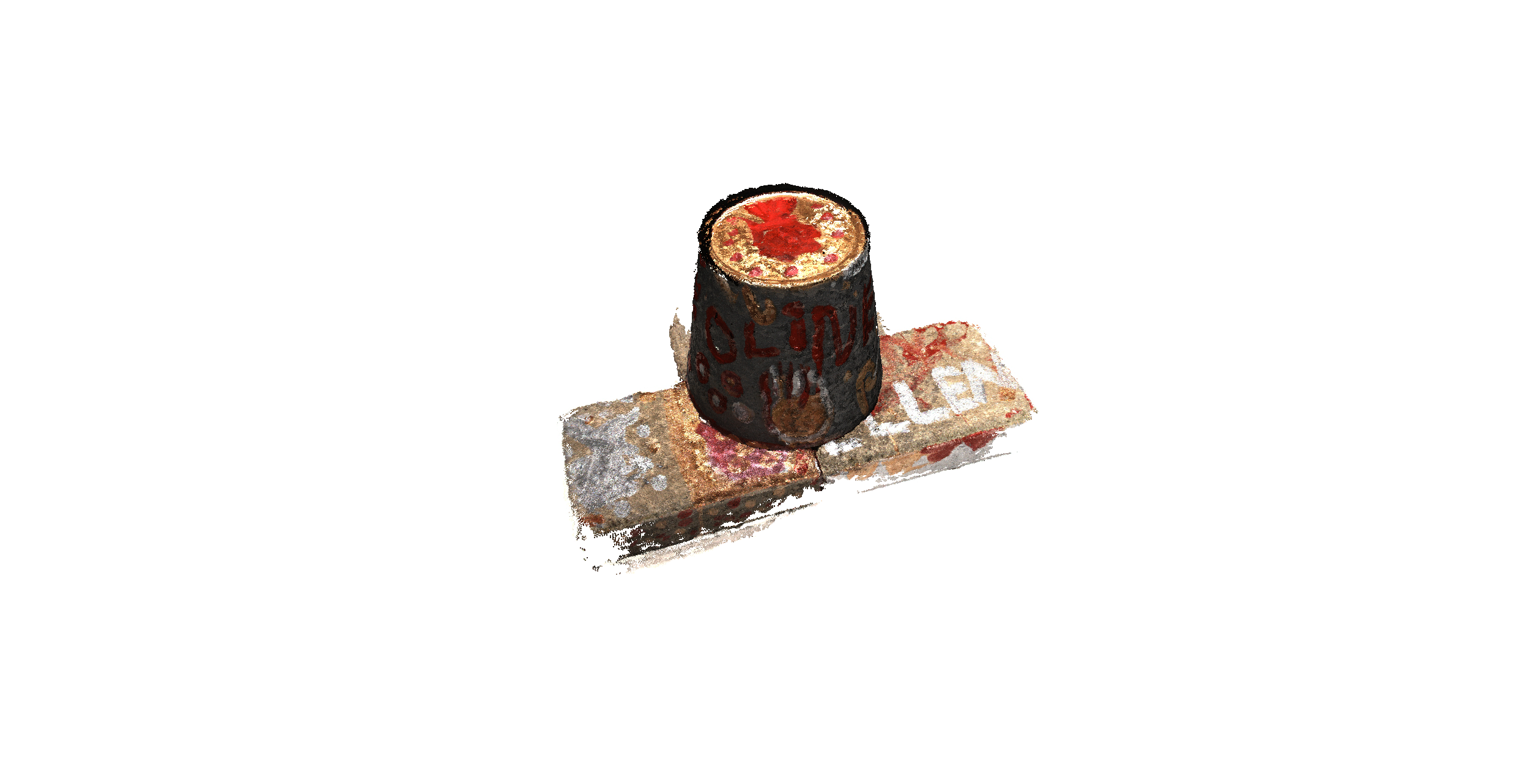}  \\
    \hline
     \includegraphics[trim={21cm 12cm 21cm 10cm},clip,width=.3\textwidth,height = 0.18\textwidth]{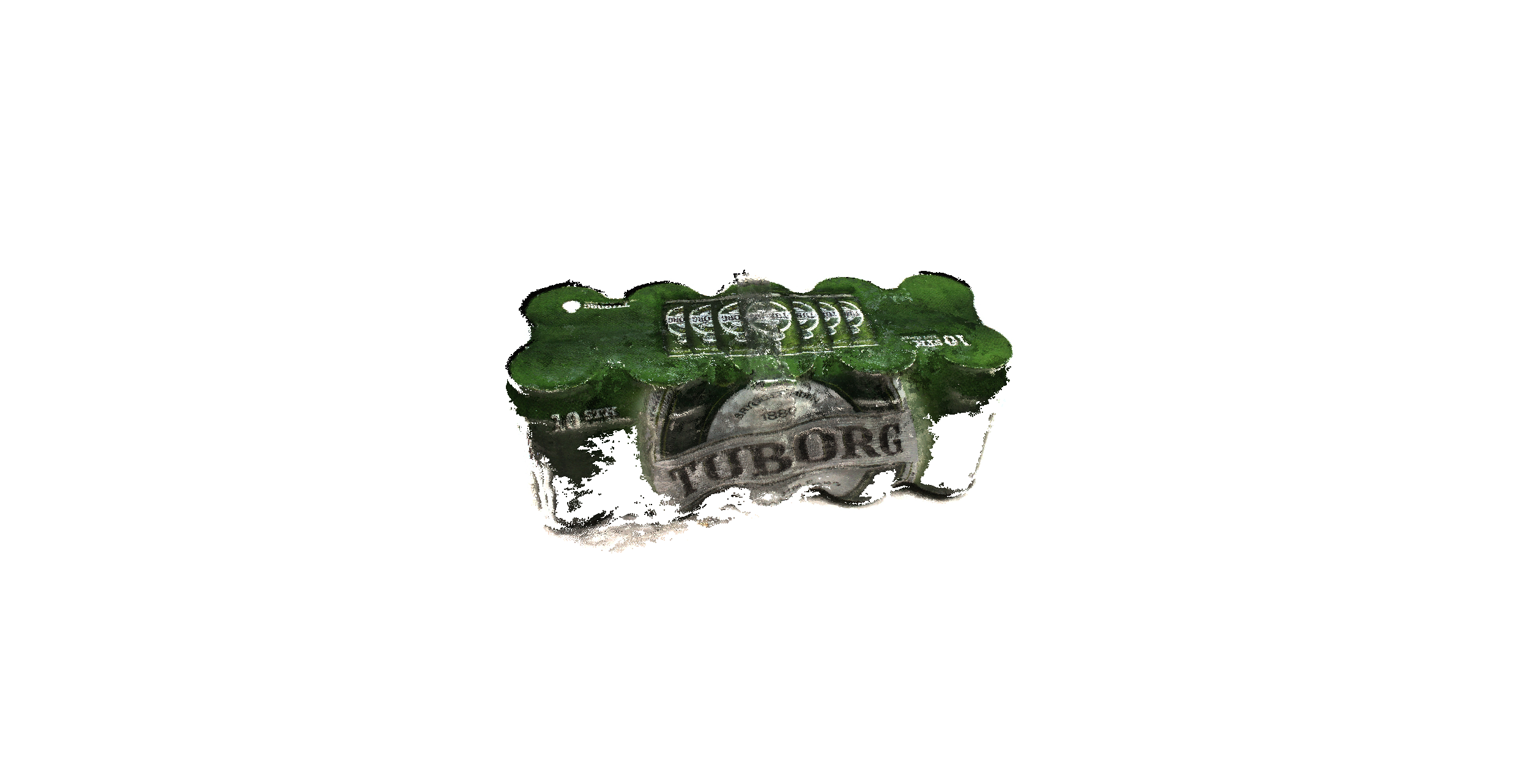} & 
    \includegraphics[trim={22cm 9cm 22cm 10cm},clip,width=.3\textwidth,height = 0.18\textwidth]{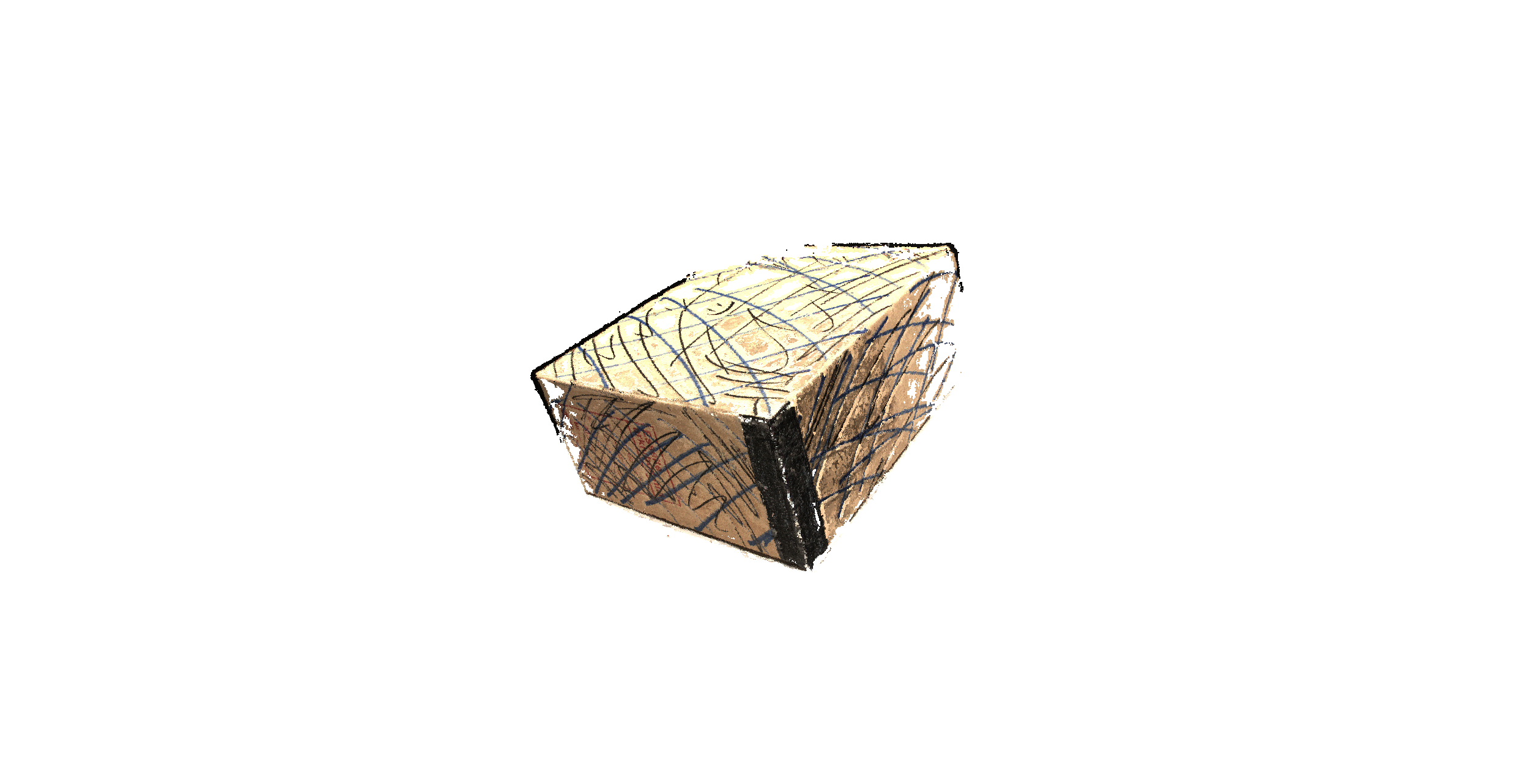}  \\
    
    \hline
    
\end{tabular}
}
\caption{Point-cloud reconstruction of DTU evaluation scans using our approach. }
\label{tbl:depth}
\end{figure}

\begin{figure}
\centering
\resizebox{\columnwidth}{!}{
\begin{tabular}{cccc}
   
     reference frame & Self DTU & Self PT bMVS, Self FT DTU & Ours (Meta, Self) \\
    
    \includegraphics[width=.25\textwidth]{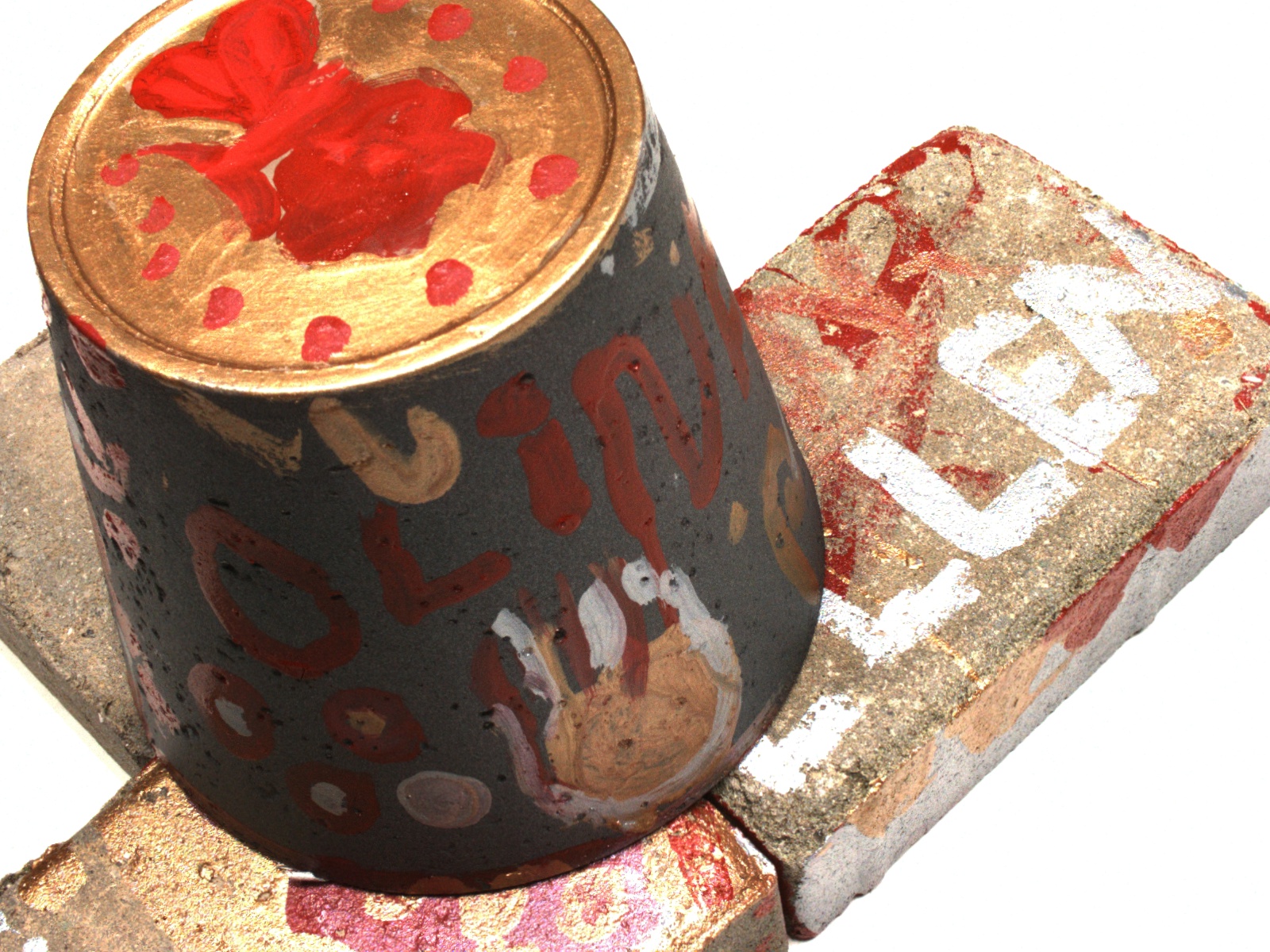} & 
    \includegraphics[width=.25\textwidth]{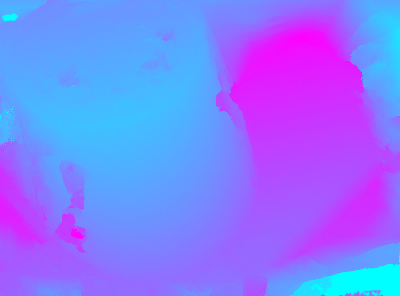} & 
    \includegraphics[width=.25\textwidth]{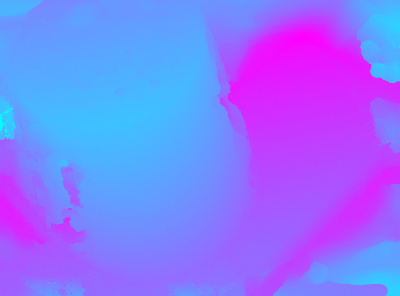} & 
    \includegraphics[width=.25\textwidth]{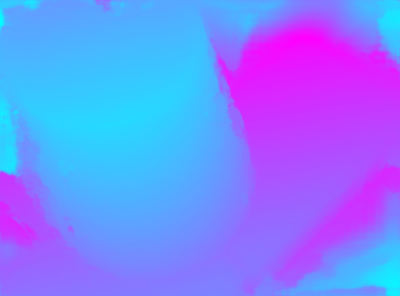} \\
    
    \includegraphics[width=.25\textwidth]{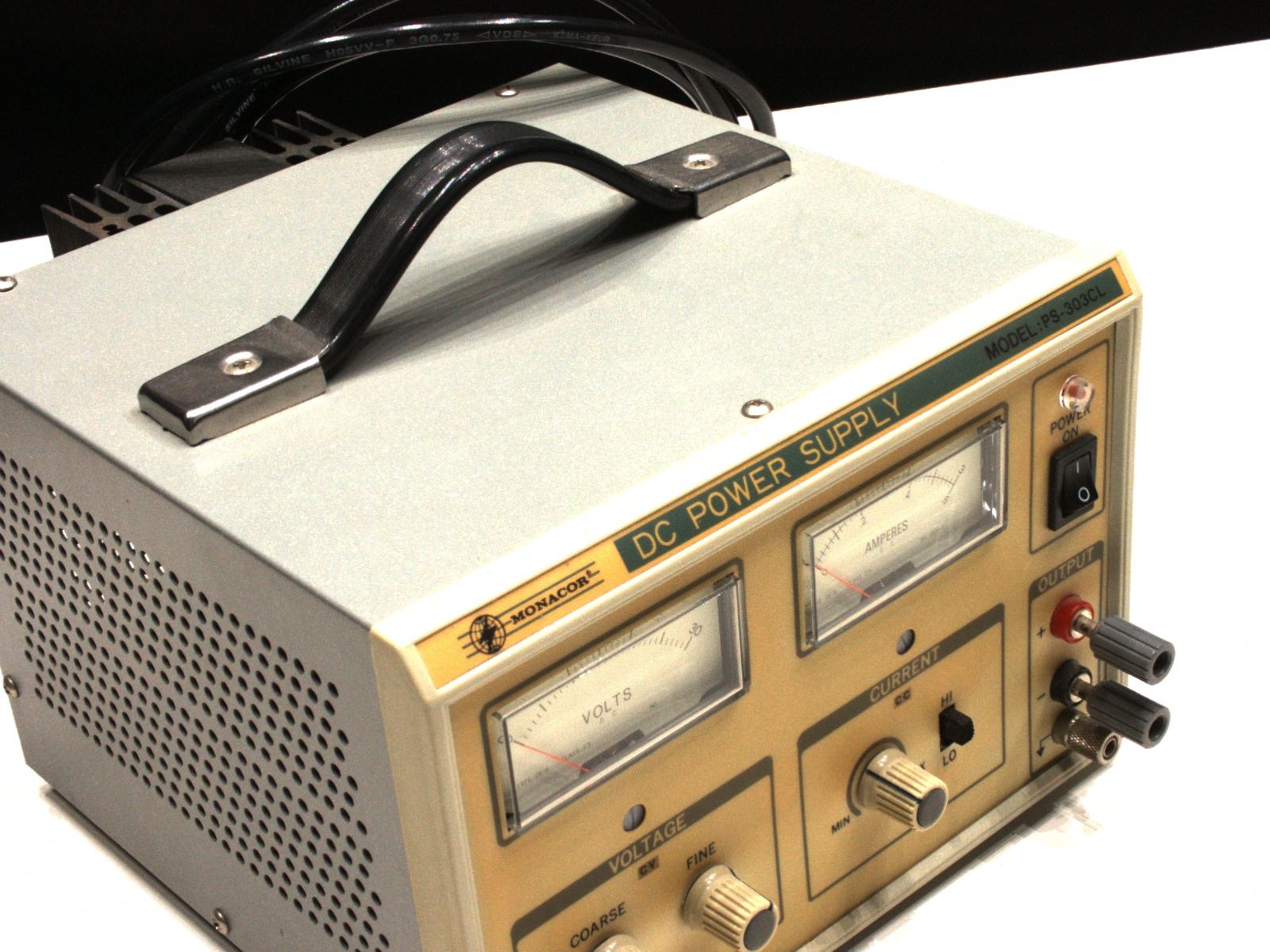} & 
    \includegraphics[width=.25\textwidth]{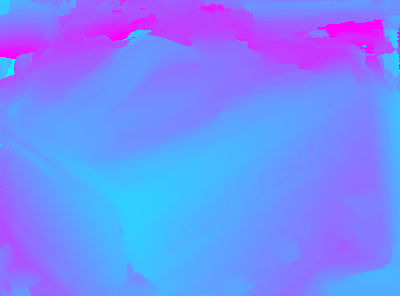} & 
    \includegraphics[width=.25\textwidth]{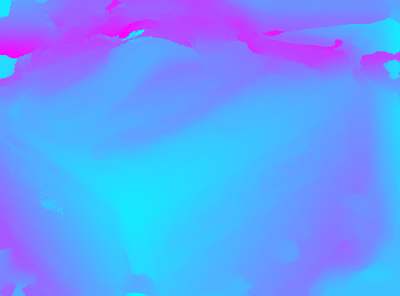} & 
    \includegraphics[width=.25\textwidth]{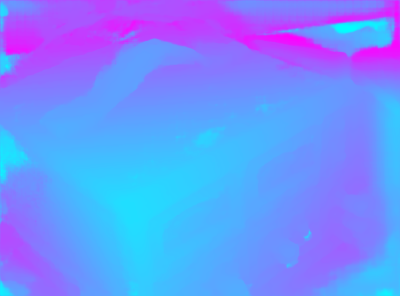} \\
    
    \includegraphics[width=.25\textwidth]{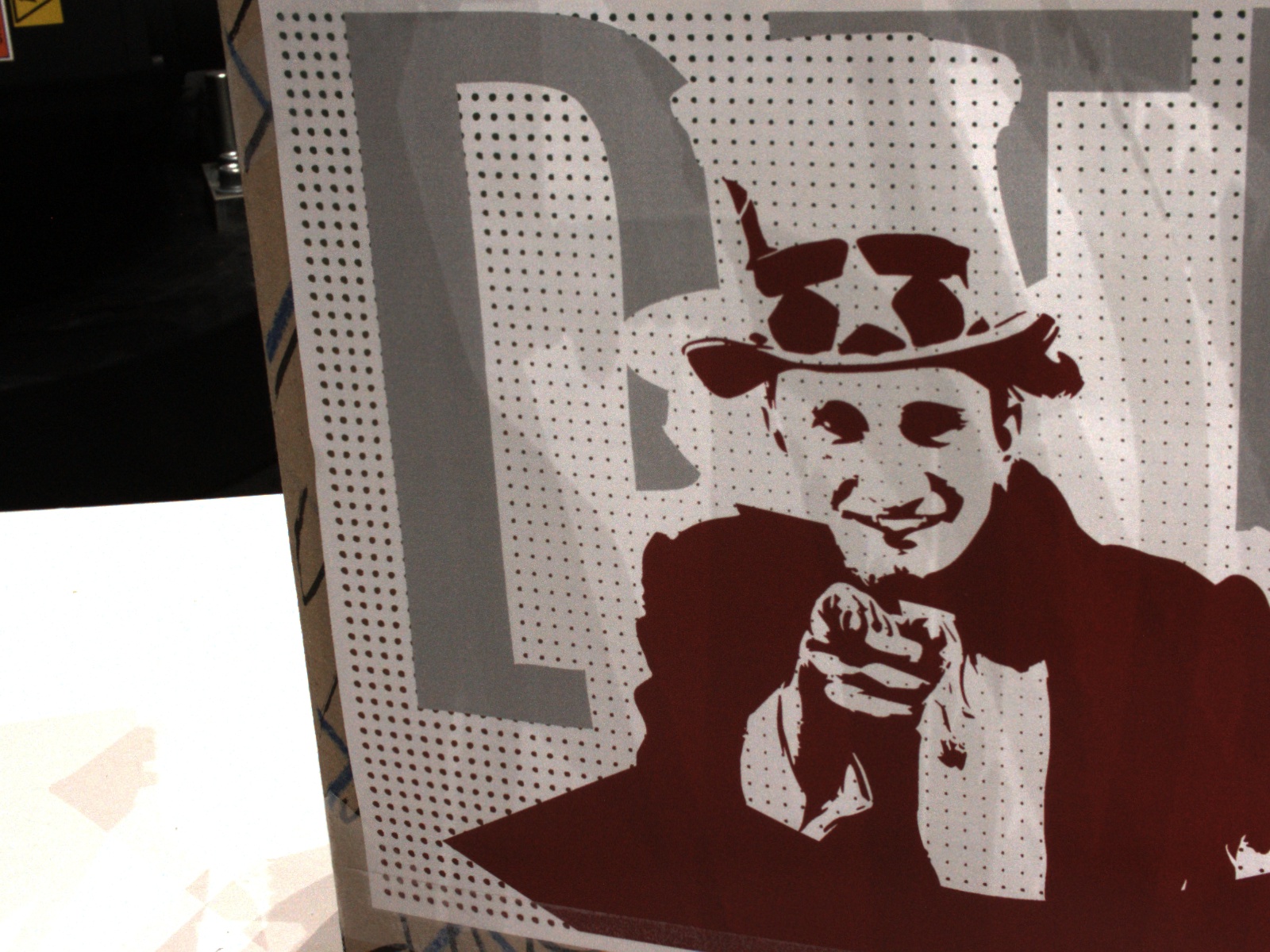} & 
    \includegraphics[width=.25\textwidth]{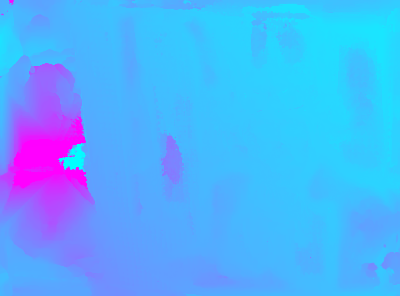} & 
    \includegraphics[width=.25\textwidth]{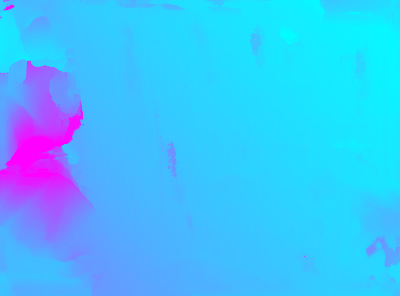} & 
    \includegraphics[width=.25\textwidth]{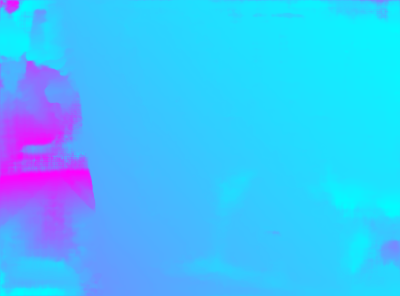} \\
    
    \includegraphics[width=.25\textwidth]{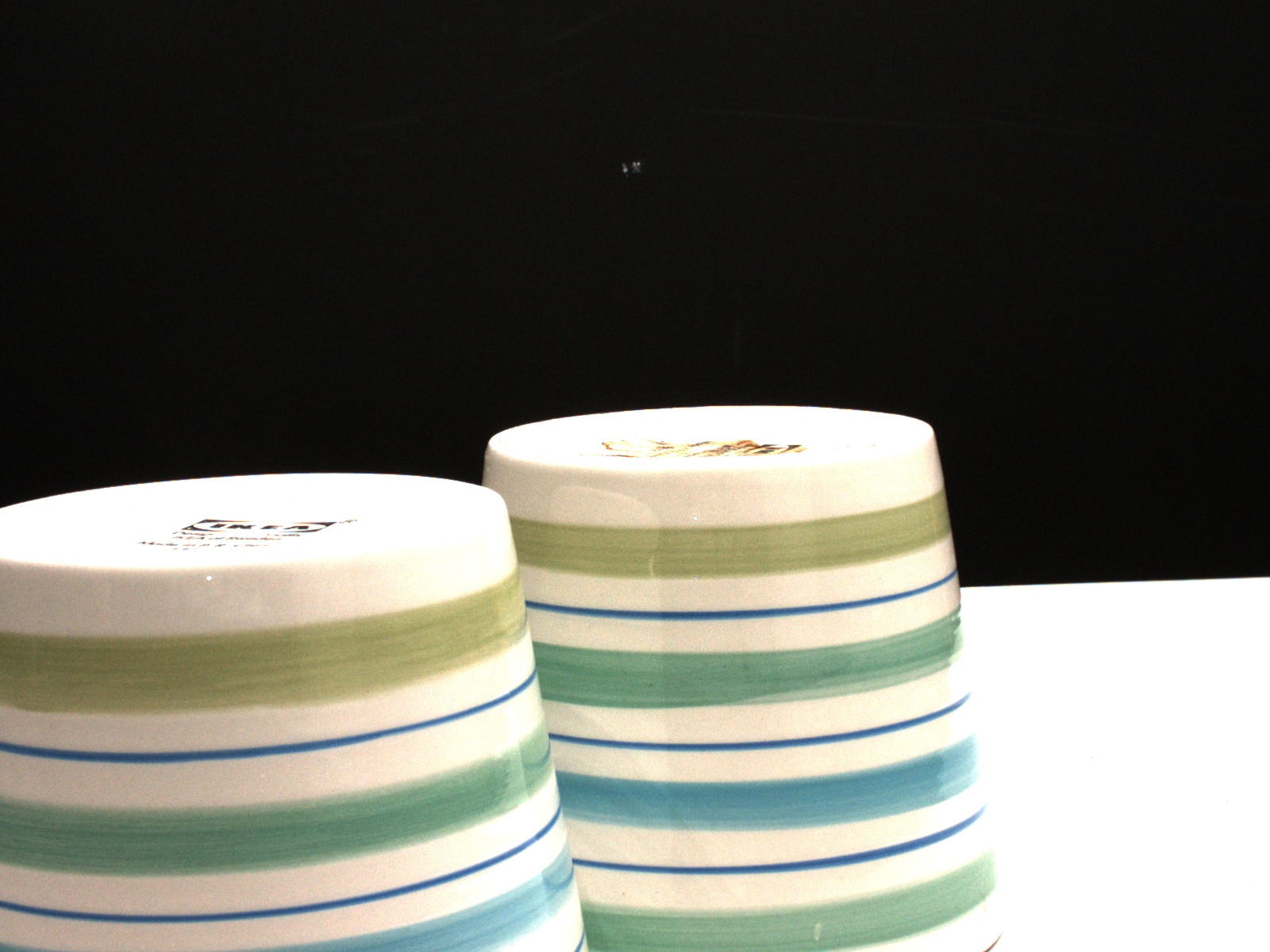} & 
    \includegraphics[width=.25\textwidth]{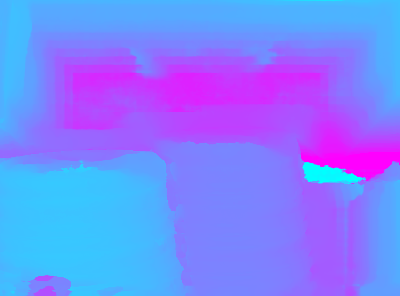} & 
    \includegraphics[width=.25\textwidth]{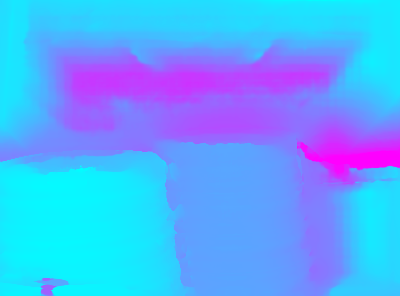} & 
    \includegraphics[width=.25\textwidth]{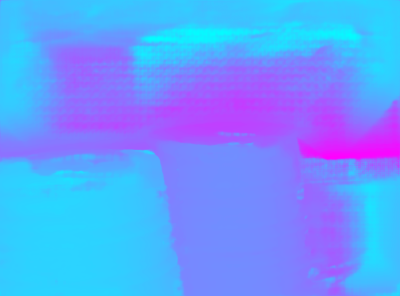} \\

\end{tabular}
}
\caption{Depth maps predicted on the DTU test set. From left to right: reference frame, depth maps predicted by our network trained self-supervised on DTU only, depth maps predicted by our network pretrained self-supervised on BlendedMVS and fine-tuned self-supervised on DTU, our meta-learning approach pretrained on BlendedMVS and fine-tuned on DTU. Our approach predicts smoother depth maps at homogeneous surfaces and provides better completeness. }
\label{tbl:depth}
\end{figure}



\end{document}